%% file: neurips_2023.tex
\documentclass{article}

\usepackage[preprint,nonatbib]{neurips_2023}

\usepackage{microtype}
\usepackage{graphicx}
\usepackage{booktabs} 

\usepackage{wrapfig}
\usepackage{hyperref}
\usepackage{algorithm}
\usepackage{algorithmic}

\usepackage{amsmath}
\usepackage{amssymb}
\usepackage{mathtools}
\usepackage{amsthm}
\usepackage{caption}
\usepackage{subcaption}

\usepackage{tikz}
\usepackage{mathptmx} 
\usepackage{fancyhdr}
\usepackage[normalem]{ulem}
\usepackage{microtype}                      
\usepackage{amssymb, amsmath, acronym}
\usepackage{multirow}
\usepackage{array}
\usepackage{tabularx}
\usepackage{graphicx}
\usepackage{arydshln}
\usepackage{booktabs}
\usepackage{csquotes}
\usepackage{pgfplots}
\usepackage{pgfplotstable}
\usepackage{wrapfig}

\definecolor{Paired-2}{RGB}{166,206,227}
\definecolor{Paired-1}{RGB}{31,120,180}
\definecolor{Paired-4}{RGB}{178,223,138}
\definecolor{Paired-3}{RGB}{51,160,44}
\definecolor{Paired-6}{RGB}{251,154,153}
\definecolor{Paired-5}{RGB}{227,26,28}
\definecolor{Paired-8}{RGB}{253,191,111}
\definecolor{Paired-7}{RGB}{255,127,0}
\definecolor{Paired-10}{RGB}{202,178,214}
\definecolor{Paired-9}{RGB}{106,61,154}
\definecolor{Paired-12}{RGB}{255,255,153}
\definecolor{Paired-11}{RGB}{177,89,40}

\definecolor{dateblue}{RGB}{36,80,117}
\definecolor{datemagenta}{RGB}{183,50,101}
\definecolor{dateorange}{RGB}{198,84,54}
\definecolor{datebrown}{RGB}{198,140,54}
\definecolor{dateyellow}{RGB}{198,178,54}

\usepackage{tikz, pgfplots}
\usetikzlibrary{spy}
\usepackage{tikzsymbols}
\usetikzlibrary{automata,positioning}
\usetikzlibrary{matrix, chains, patterns}
\usetikzlibrary{decorations.pathreplacing,calligraphy,shapes}

\pgfdeclarelayer{background}
    \pgfdeclarelayer{foreground}
    \pgfsetlayers{background,main,foreground}
\usepackage{fontawesome}
\usepackage[capitalize,noabbrev]{cleveref}

\newif\ifshowcomments
\showcommentstrue
\ifshowcomments
    \newcommand{\altan}[1]{\textbf{\color{orange}[Altan: #1]}}
    \providecommand{\alvin}[1]{{\protect\color{purple}{\bf [alvin: #1]}}}
\else
    \newcommand{\altan}[1]{}
    \providecommand{\alvin}[1]{}
\fi

\ifshowcomments
    \newcommand{\ksen}[1]{\textbf{\color{red}[ksen: #1]}}
\else
    \newcommand{\ksen}[1]{}
\fi

\theoremstyle{plain}

\theoremstyle{definition}

\theoremstyle{remark}

\usepackage[textsize=tiny]{todonotes}

\usepackage[utf8]{inputenc} 
\usepackage[T1]{fontenc}    
\usepackage{hyperref}       
\usepackage{url}            
\usepackage{booktabs}       
\usepackage{amsfonts}       
\usepackage{nicefrac}       
\usepackage{microtype}      
\usepackage{xcolor}         

\title{{\sc \textbf{SlimFit}}: Memory-Efficient Fine-Tuning of Transformer-based Models Using Training Dynamics}

\author{ 
  Arash Ardakani$^{1}$ \quad Altan Haan$^{1}$ \quad Shangyin Tan$^{1}$ \quad Doru Thom Popovici$^{2}$ \\ \textbf{Alvin Cheung}$^{1}$ \quad   \textbf{Costin Iancu}$^{2}$ \quad \textbf{Koushik Sen}$^{1}$ \\
  University of California, Berkeley$^{1}$ \quad Lawrence Berkeley National Laboratory$^{2}$\\
  \texttt{\{arash.ardakani,altanh,shangyin,akcheung,ksen\}@berkeley.edu}  \\
  \texttt{\{dtpopovici,cciancu\}@lbl.gov}
}

\begin{document}

\maketitle

\vspace{-0.75cm}
\begin{abstract}
\vspace{-0.25cm}
  Transformer-based models, such as BERT and ViT, have achieved state-of-the-art results across different natural language processing (NLP) and computer vision (CV) tasks. However, these models are extremely memory intensive during their fine-tuning process, making them difficult to deploy on GPUs with limited memory resources. To address this issue, we introduce a new tool called {\sc SlimFit} that reduces the memory requirements of these models by dynamically analyzing their training dynamics and freezing less-contributory layers during fine-tuning. The layers to freeze are chosen using a runtime inter-layer scheduling algorithm. {\sc SlimFit} adopts quantization and pruning for particular layers to balance the load of dynamic activations and to minimize the memory footprint of static activations, where static activations refer to those that cannot be discarded regardless of freezing. This allows {\sc SlimFit} to freeze up to $95\%$ of layers and reduce the overall on-device GPU memory usage of transformer-based models such as ViT and BERT by an average of $2.2\times$, across different NLP and CV benchmarks/datasets such as GLUE, SQuAD 2.0, CIFAR-10, CIFAR-100 and ImageNet with an average degradation of $0.2\%$ in accuracy. For such NLP and CV tasks, {\sc SlimFit} can reduce up to $3.1\times$ the total on-device memory usage with an accuracy degradation of only up to $0.4\%$. As a result, while fine-tuning of ViT on ImageNet and BERT on SQuAD 2.0 with a batch size of 128 requires 3 and 2 32GB GPUs respectively, {\sc SlimFit} enables their fine-tuning on a single 32GB GPU without any significant accuracy degradation. The code of this paper is available at \url{https://github.com/arashardakani/SlimFit}.
\end{abstract}

\vspace{-0.75cm}
\section{Introduction}
\vspace{-0.25cm}
\label{intro}
Over the past few years, various transformer-based models have been developed with the adoption of the attention mechanism that weighs the importance of each part of the input data differently. Pre-training of such transformer-based models on large data has led to a significant boost in accuracy when fine-tuned on various natural language processing (NLP) and computer vision (CV) downstream tasks \cite{BERT,ViT}. Despite their great performance in achieving state-of-the-art (SOTA) accuracy, these models are memory intensive and require a considerably large amount of on-device GPU memory during their fine-tuning phase when compared to the conventional convolutional and recurrent neural networks \cite{checkmate}. The memory requirement of current transformer-based models has made them difficult to fine-tune even on powerful GPUs. With the introduction of larger transformer-based models over the past few years, the on-device GPU memory has become a major bottleneck for their fine-tuning process \cite{checkmate, GACT, DropIT}.

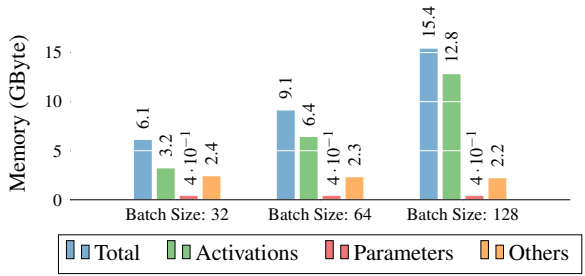
\begin{wrapfigure}{r}{0.6\textwidth}
\center
\vspace{-0.65cm}
\scalebox{0.95}{
\input{fig1.tex}}
\caption{The breakdown of memory usage of BERT when fine-tuned on different batch sizes including 32, 64, and 128.}
    \label{fig1}
    \vspace{-0.25cm}
\end{wrapfigure}

The total on-device memory usage of GPUs consists primarily of activations, parameters, gradients, optimizer states, and the CUDA context. Among these factors, activations account for most of the memory usage due to batching, which makes them several orders of magnitude larger than other factors (see Fig. \ref{fig1}). Therefore, activation compressed training (ACT) has emerged as the primary solution for memory-efficient fine-tuning \cite{ACT, GACT}. This approach first compresses activations during the forward pass and then decompresses them during the backward pass. In this way, the memory footprint can be
significantly reduced by caching the compressed activations. In ACT, quantization \cite{Approx_ACT, TinyScript, ACT, GACT} has been a popular choice to compress activations among other compressors such as JPEG \cite{JPEG} or pruning \cite{DropIT}. The current SOTA ACT adaptively assigns quantization bits to each layer for a given architecture \cite{GACT}. While the SOTA ACT successfully reduces the memory footprint of activations, its overall on-device GPU memory reduction is not significant. For instance, the total on-device GPU memory reduction of the SOTA ACT is limited to 0.1GB despite its $6.4\times$ reduction in the memory of activations when fine-tuning BERT on CoLA dataset with a batch size of 32. It is worth mentioning that we refer to the memory usage reported by ``nvidia-smi'' as the overall on-device memory in this paper (see Appendix \ref{mem_management} for more information on memory management).

Tensor rematerialization \cite{checkmate, checkpoint, Remat_offload, dtr}, also known as gradient checkpointing, is another prominent approach to reducing activation memory by trading computations for memory. In tensor rematerialization, only specific activations are stored during the forward pass, while the rest are recomputed in the backward pass. Of course, recomputing activations requires more operations and significantly prolongs the fine-tuning process \cite{GACT}. Reduced precision training, as another approach, performs the computations of both forward and backward passes in low-precision \cite{MPT, TII, 8FP, SM8T}. While these works can successfully train conventional models, few-bit model fine-tuning is not trivial. For instance, 8-bit quantization of BERT for inference results in a significant precision loss \cite{Q8BERT}, which makes fine-tuning on few bits a challenging task.

Low-rank adaptation (LoRA) \cite{lora} is another key approach to reducing the overall on-device GPU memory where the transformer-based models are fine-tuned by inserting a small number of trainable parameters into each layer while keeping the pre-trained model parameters frozen. Such an approach enables fine-tuning transformer-based models with significantly less number of trainable parameters, leading to a reduction in the memory footprint of optimizer states and gradients. Such a memory reduction becomes significant for extremely large transformer models such as GPT \cite{gpt} with over hundred billion parameters.


Different from these methods, we put forward a new approach to reducing the overall on-device memory usage by analyzing training dynamics. More precisely, we dynamically analyze the gradient contributions of layers in transformer-based models and perform parameter updates for specific layers only while the rest of layers are kept frozen. Training dynamics have been used to analyze the behavior of a model during its training/fine-tuning process \cite{DC, ESTD, TDANN}. However, our work uses training dynamics to detect and discard unimportant activations during fine-tuning by freezing their associated layers, leading to a reduction of the memory footprint. Our method is orthogonal to existing approaches including rematerialization and LoRA, which could be composed for further reductions. 

Freezing layers or parameters has been studied in different domains, including transformer-based models to preserve previously learned information during fine-tuning \cite{SFT}. Freezing parameters have also been used to regularize fine-tuning (e.g., over-fitting reduction) in pre-trained models \cite{ACF}. Recently, freezing has been used to accelerate fine-tuning by progressively freezing model blocks \cite{AutoFreeze, smartfrz, pipetrans}. However, since such an approach starts the fine-tuning process without freezing at least for a few training iterations, its overall on-device memory requirement remains similar to that of training without freezing. For instance, fine-tuning ViT on ImageNet with a batch size of 128 using such a freezing approach on a single 32GB GPU results in an out-of-memory error (see Appendix \ref{Freezing_comp} for more details).

To orchestrate effective layer-freezing decisions, we introduce a runtime inter-layer scheduling (ILS) algorithm. Our method finds and freezes a set of layers in transformer-based models that are less contributory, i.e., layers with fewer updates in their parameters, to the fine-tuning process at each iteration. While the ILS algorithm successfully detects and freezes unimportant layers, its memory reduction is not proportional to the freezing rate. The reason behind this disproportionality is twofold: the imbalanced number of activations among layers and the existence of static activations. Static activations refer to those that cannot be discarded regardless of freezing (e.g., activations of non-linear functions such as GELU). We address these two issues using quantization and pruning to even out the number of activations across all layers and to reduce the memory overhead of static activations. We use quantization and pruning for a few specific layers of transformer-based models as opposed to reduced precision training methods where all the layers are quantized. As a result, the impact of quantization and pruning on accuracy is insignificant in our work. For instance, the accuracy degradation due to quantization and pruning is only $0.1\%$ on the MRPC dataset.

By combining ILS with quantization and pruning, we introduce a performance tool called {\sc SlimFit} for reducing the on-device GPU memory usage of transformer-based models during fine-tuning. We demonstrate the effectiveness of {\sc SlimFit} in reducing the memory footprint on popular models of BERT and ViT. We show that {\sc SlimFit} can freeze up to $95\%$ of layers and reduce the overall on-device memory usage by an average of $2.2\times$ when fine-tuning BERT and ViT models on different benchmarks and datasets, such as GLUE, SQuAD 2.0, CIFAR-10, CIFAR-100 and ImageNet with an average accuracy degradation of $0.2\%$. More precisely, {\sc SlimFit} reduces the overall on-device memory usage of the fine-tuning process on GLUE from 6.1GB to 4.0GB ($1.5\times$ reduction) with a batch size of 32, on SQuAD 2.0 from 58.5GB to 19.1GB ($3.1\times$ reduction) with a batch size of 128, on CIFAR-10 from 7.2GB to 4.3GB ($1.7\times$ reduction) with a batch size of 32, on CIFAR-100 from 7.2GB to 4.5GB ($1.6\times$ reduction) with a batch size of 32, and on ImageNet from 77.4GB to 26.1GB ($3.0\times$) with a batch size of 128 at the cost of up to $0.4\%$ accuracy degradation. As a result, {\sc SlimFit} enables performing memory-intensive fine-tuning processes on a single 32GB GPU such as fine-tuning ViT on ImageNet with a batch size of 128 while this  normally requires three 32GB GPUs. 

\vspace{-0.25cm}
\section{Preliminaries}
\vspace{-0.25cm}
Over the past few years, pre-training of attention-based models has led to significant advances on many NLP and CV tasks with the popular BERT \cite{BERT} and ViT \cite{ViT} models. The pre-training process provides a good initialization point such that these models can better generalize on unseen data of downstream tasks. Therefore, these models can achieve state-of-the-art results by fine-tuning through small adjustments to their parameters. 
Architecturally, these models consist of an initial embedding layer, followed by repeated blocks of multi-head attention (MHA) fed into a feed-forward network (FFN) module (see Appendix \ref{Arch} for more details). The base architectures of BERT and ViT contain over a hundred layers built up in this manner.

Despite the large number of layers, not all need to be updated during fine-tuning to achieve decent performance on downstream tasks, as shown in \cite{What_happens}. Notably, the authors found that freezing approximately $60\%$ of early attention layers in BERT led to negligible performance degradation. This suggests that the fine-tuned model tends to preserve generic features learned during pre-training. Motivated by this study, we seek to analyze the training dynamics of pre-trained models and to automatically detect layers with less contributions to the fine-tuning process.

\vspace{-0.35cm}
\section{Learning the Importance of Layers} \label{ILS_sec}
\vspace{-0.25cm}
Training dynamics is an active field of research that provides insight about the behavior of pre-trained models when fine-tuning on downstream tasks. The convergence proof of optimization algorithms such as stochastic gradient descent \cite{SGDproof} shows that the distance between the parameters and the optimal solution is reduced over training iterations and accordingly, the weight distance (or the weight update amount) between consecutive iterations decreases. Therefore, it is possible that some layers can only receive minimal changes to their parameters as we approach the end of the training process. Of course, detecting and freezing such layers, when they show minimal updates, will not affect accuracy. Since transformer-based models are pre-trained, they already show small updates during fine-tuning compared to pre-training. As such, detecting and freezing layers with minimal updates (i.e., weight distance values) will not significantly affect the fine-tuning process and accordingly the final accuracy. Based on the above observations, we consider the $\ell_1$-norm of the update received by parameters of each layer through all the fine-tuning iterations as the training dynamics in this paper. It is also worth mentioning that freezing layers has no impact on training convergence as it causes a pause in the training procedure of frozen layers as shown by our theoretical analysis in Appendix \ref{conv_analysis}. 

\vspace{-0.35cm}
\subsection{Training Dynamics}
\vspace{-0.25cm}
Let us consider a pre-trained model with a set of parameters $\mathbf{W}$ where the parameters associated with the $i$th layer at iteration $t$ is denoted as $\mathbf{W}_i^t \in \mathbb{R}^{M \times I}$. The training dynamics of for the $i$th layer at iteration $t$ is defined as the $\ell_1$-norm of the distance between $\mathbf{W}_i^{t-1}$ and $\mathbf{W}_i^{t}$, i.e., 
\vspace{-0.1cm}
\begin{equation} 
d_i^t = \dfrac{1}{M \times I}\left\lVert\dfrac{W_i^{t} - W_i^{t-1}}{W_i^{t-1}} \right\rVert_{\ell_1},
\label{TD}
\end{equation}
where $\mathbf{d}^t \in \mathbb{R}_+^n$ containing all $d_i$s at iteration $t$ is referred to as distance vector, and $n$ denotes the total number of layers. In fact, Eq. (\ref{TD}) calculates the normalized change in the parameters of the $i$th layer.

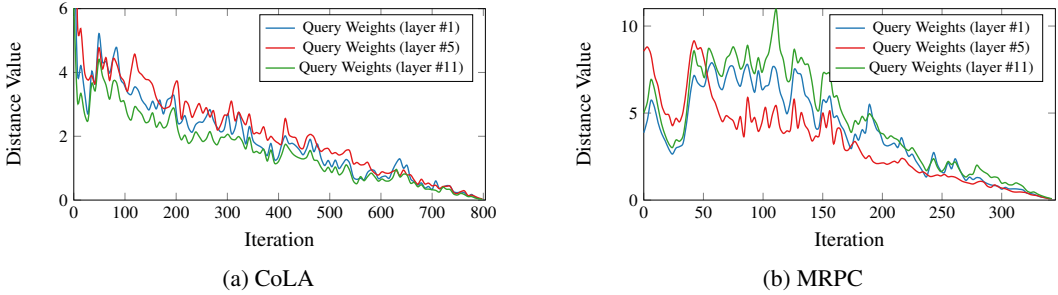
\begin{figure}[!t]
\center
\pgfplotsset{width=10cm, height = 5.5cm}
\begin{subfigure}[t]{0.5\textwidth}
\scalebox{0.65}{
\input{fig3a.tex}}
\caption{CoLA}
\label{fig3a}
    \end{subfigure}%
\hfill
    \begin{subfigure}[t]{0.5\textwidth}
        \center
        \scalebox{0.65}{
        \input{fig3b.tex}}
\caption{MRPC}
\label{fig3b}
    \end{subfigure}%
\caption{The distance values of query weight matrix for the first, fifth and eleventh attention layers of BERT-base fine-tuned on (a) CoLA  and (b) MRPC datasets for 3 epochs.}
\label{fig3}
\vspace{-0.65cm}
\end{figure}

\vspace{-0.25cm}
\subsection{Inter-Layer Scheduling Algorithm} \label{ILSM}
\vspace{-0.25cm}
We use the distance values as training dynamics to analyze the fine-tuning behavior of pre-trained models. For instance, consider the distance values across all the fine-tuning iterations for the CoLA \cite{CoLA} and MRPC \cite{GLUE} datasets. Fig. \ref{fig3a} shows the distance values of the query weight matrix for the first, fifth and eleventh attention layers of BERT-base fine-tuned on CoLA dataset whereas Fig. \ref{fig3b} depicts those of the same layers for BERT-based fine-tuned on MRPC dataset. 

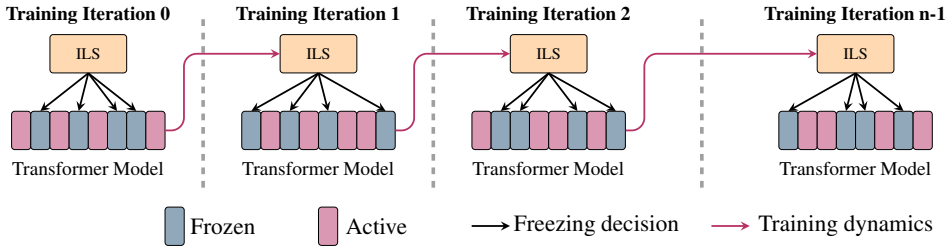
\begin{figure}[b]
\vspace{-0.3cm}
\center
\scalebox{0.85}{
\input{fig2_1.tex}}
\caption{The overview of ILS algorithm. ILS freezes a certain number of layers depending on the freezing rate at every single iteration throughout the fine-tuning process for the total of $n$ training iterations.}
    \label{fig2}
\vspace{-0.2cm}
\end{figure}

We observe the following based on the experimental results of these two datasets. First, the updated amount for each layer becomes smaller over fine-tuning iterations. Second, the updated amount of each layer is task-specific and is independent of its position. Third, there are some layers showing smaller distance values w.r.t. other layers across almost all the iterations.
Finally, layers with a higher distance value in the beginning can become smaller over the fine-tuning iterations than layers starting with a lower distance value.

Given the above observations, we introduce an ILS algorithm to decide on updating priority of layers using their distance values. Fig. \ref{fig2} shows the overview of the ILS algorithm. At each iteration ranging from the first iteration to the last iteration, our ILS algorithm selects those layers with large distance values to be updated and those with small distance values to be frozen. More precisely, layers are first ranked based on their distance values at each training iteration and then those with small distance values are kept frozen according to the freezing rate as a hyper-parameter. The intuition is that layers with small distance values are less contributory to the fine-tuning process as their parameters are not being updated much. On the other hand, the layers with large distance values are learning task-specific patterns by making more significant adjustments to their parameters. Note that freezing middle layers does not interrupt the gradient propagation to the early layers of the network as shown through an example in Appendix \ref{backprop_frozen_layer}.

The freezing rate of the ILS algorithm can be decided based on the on-device GPU memory budget. Of course, using an extremely high freezing rate may result in a performance degradation depending 
\begin{minipage}[t]{\textwidth}
  \begin{minipage}[b]{0.6\textwidth}
    \centering
    \scalebox{0.53}{
\input{fig4.tex}}
    
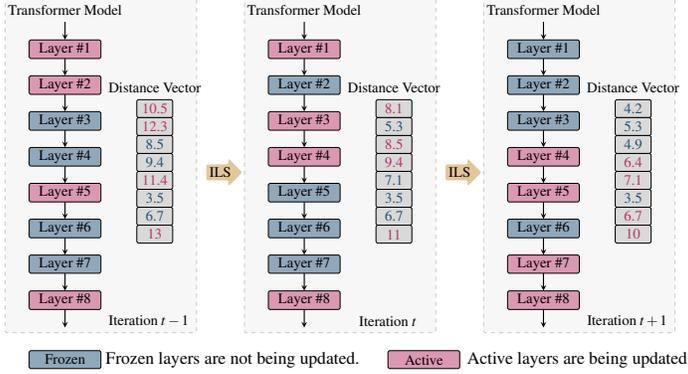
\captionof{figure}{An example of the iterative freezing process using our ILS algorithm.}
    \label{fig4}
  \end{minipage}
  \hfill
    \scalebox{0.85}{\begin{minipage}{0.4\textwidth}
    \vspace{-8cm}
      \begin{algorithm}[H]
        \caption{The pseudo code of the ILS algorithm performing iterative freezing.}
        \label{alg1}
        \begin{algorithmic}
   \STATE {\bfseries Input:} model, number of iterations as $itr$, number of layers as $L$, freezing rate $F$
   
   \STATE $d$ = rand($L$)
   
   \FOR{$i$ {\bfseries in} $itr$}
   
   \STATE $idx$ = argsort($d$)[:int(L*F)]
   
   \FOR{$j$ {\bfseries in} $idx$}
   
   \STATE model.layer[$j$].requires\_grad = False
   
   \ENDFOR
   
   \STATE model.train()
   
   \STATE Update $d$
   
   \ENDFOR
\end{algorithmic}
      \end{algorithm}
    \end{minipage}}
    \end{minipage}
on the downstream task, providing a worthwhile trade-off between accuracy and on-device GPU memory. On the other hand, while performance degradation is unlikely with a very small freezing rate, the memory reduction is insignificant as well.

Since there is no prior knowledge about the distance values of each layer at the beginning of the fine-tuning process, our ILS algorithm initializes the distance vector with large random values. Depending on the freezing rate, each layer along with its distance value are updated during the first few iterations once until all random numbers in the distance vector are substituted with an actual distance value. Afterwards, layers are kept frozen according to their actual distance value. The distance value of the active layers is only updated at each iteration while that of the frozen layers remains unchanged. The pseudo code of our ILS algorithm performing iterative freezing is shown in Algorithm \ref{alg1}.


To better understand the ILS algorithm, we illustrate the iterative freezing process using an example as shown in Fig. \ref{fig4}. Suppose we have an 8-layer transformer-based model and accordingly an 8-element distance vector at iteration $t$. Considering the freezing rate of $50\%$ for this example, 4 layers with the lowest distance values are kept frozen and the rest are updated at each iteration. 

\vspace{-0.3cm}
\section{Inter-Layer Load-Balancing} \label{ILLB_sec}
\vspace{-0.25cm}
So far, we have introduced our ILS algorithm that prioritizes updating particular layers while keeping the rest of layers frozen according to their distance value. For the given freezing rate of $50\%$ as an example, we expect to see a $2\times$ reduction in the memory footprint of activations. However, this is not the case in transformer-based models due to the imbalanced the number of activations across all the layers. In fact, the imbalance in the number of activations undermines the ability of our ILS algorithm in reducing the memory footprint during the fine-tuning as shown in Fig. \ref{fig5}.

Since the focus of this paper is on transformer-based models such as BERT and ViT, we analyze their architecture for imbalanced layers. Table \ref{tab1} summarizes the number of activations associated to the input of layers with trainable parameters in BERT or ViT. Among all trainable layers, there is only one imbalanced layer in the attention block which contains $4\times$ more activations than other layers.

To address the load-balancing issue in the number of activations for the aforementioned layer, we use quantization. Since the imbalance factor among layers is $4\times$, 
we adopt 8-bit quantization for activations of the imbalanced layer where 4 bits are used for both the integer and fractional parts. In this way, the memory cost of the activations are evened out using quantization. In our quantization scheme, we cache the activations of the imbalanced layer using 8 bits during the forward pass. In the backward pass, we convert the 8-bit activations to 32-bit floating-point format. Therefore, all the forward and backward computations are still performed using single-precision floating-point format. The conversion process between 8-bit fixed-point and 32-bit floating-point formats are provided in Appendix \ref{App_conv}.

\vspace{-0.35cm}
\section{Dynamic and Static Activations} \label{SA_sec}
\vspace{-0.25cm}
The type of activations in transformer-based models can be divided into two categories: dynamic and static. We refer to the activations that can be discarded by freezing their layer as dynamic activations. On the other hand, static activations cannot be discarded regardless of freezing. Among different
\begin{minipage}[t]{\textwidth}
  \begin{minipage}[b]{0.44\textwidth}
    \centering
    \scalebox{0.8}{
\input{fig5.tex}}
    
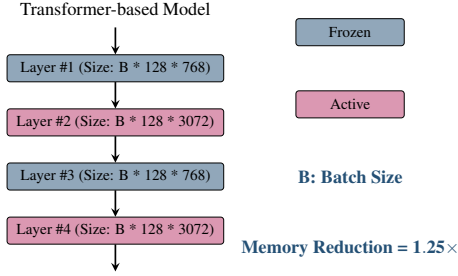
\captionof{figure}{An example of a model with imbalanced number of activations and its impact on the memory reduction.}
    \label{fig5}
  \end{minipage}
  \hfill
  \begin{minipage}[b]{0.54\textwidth}
    \centering
    \scalebox{0.75}{
\input{tab1.tex}}
      \captionof{table}{The number of activations associated to the input of layers with trainable parameters in BERT where $B$, $T$, $H$ denote the batch size, sequence length, hidden size, respectively. ViT has the same structure with different descriptions.}
      \label{tab1}
    \end{minipage}
  \end{minipage}
types of layers, GELU, MatMul, Softmax and LayerNorm contain static activations as shown Table \ref{tab2}. Note that MatMul and Softmax share the same activations. For the backward computations of Softmax, its output during the forward pass is saved as its activations. On the other hand, the input to MatMul is required for its backward computations as activations. Since the output of Softmax is an input to MatMul in the forward pass, they share the same activations.

GELU and MatMul/Softmax do not have any trainable parameters and accordingly cannot be frozen. Therefore, these two layers hold on to their activations throughout the fine-tuning process. The best approach to reduce their memory cost is quantization. We use 4 and 8 bits for quantization of activations in GELU and MatMul/Softmax, respectively. Since there is no 4-bit tensor support in PyTorch, we store each two 4-bit activations as a single 8-bit activations using shift operations. Note that using such bit-levels result in a negligible accuracy degradation while further quantization of those activations incurs a significant accuracy loss.

\begin{table}[b]
\vspace{-0.5cm}
\center
\caption{The type of activations of layers in MHA and FFN of BERT and ViT.}
\scalebox{0.75}{
\input{tab2.tex}}
\label{tab2}
\vspace{-0.25cm}
\end{table}

As opposed to GELU and MatMul/Softmax, LayerNorm contains trainable parameters and can be frozen by the ILS algorithm. However, its activations are still static. The forward pass of LayerNorm is computed by: \vspace{-0.2cm} 
\begin{equation} 
\widetilde{\mathbf{x}} = \dfrac{\textbf{x} - \mathbb{E}(\textbf{x})}{\sqrt{\text{Var}(\textbf{x}) + \epsilon}},
\end{equation} 
\begin{equation}
\textbf{y} = \widetilde{\mathbf{x}} * \mathbf{\gamma} + \mathbf{\beta},
\end{equation} 
where $\gamma$ and $\beta$ are trainable parameters. The input and output to LayerNorm are denoted by $\textbf{x} \in \mathbb{R}^H$ and $\textbf{y}\in \mathbb{R}^H$, respectively. $\mathbb{E}(\cdot)$ and $\text{Var}(\cdot)$ compute the average and variance, respectively. The derivative of the loss with respect to $\gamma$ (i.e., $\widehat{\gamma}$) is computed by
\vspace{-0.2cm} 
\begin{equation}
\widehat{\gamma} = \widetilde{\mathbf{x}} * \widehat{\mathbf{y}},
\label{LNP} \vspace{-0.2cm}
\end{equation} 
and with respect to $\beta$ (i.e., $\widehat{\beta}$) by:
\begin{equation}
\widehat{\beta} = \widehat{\mathbf{y}},
\end{equation}
where $\widehat{\mathbf{y}}$ denotes the derivative of the loss w.r.t. $y$. We also need to compute the derivative of the loss with respect to $\textbf{x}$ (i.e., $\widehat{\mathbf{x}}$) as:
\begin{equation}
\textbf{g} = \dfrac{\gamma * \widehat{\mathbf{y}}}{H * \sqrt{\text{Var}(\textbf{x}) + \epsilon}},
\end{equation}
\begin{equation}
\widehat{\mathbf{x}} = H * \textbf{g} - \sum_H \textbf{g} - \widetilde{\mathbf{x}} * \sum_H (\textbf{g} * \widetilde{\mathbf{x}}).
\label{LNI}
\end{equation}
When LayerNorm is frozen, there is no need to compute Eq. (\ref{LNP}). However, the activations of this layer cannot be discarded since they are still a part of the computations in Eq. (\ref{LNI}). More precisely, the standardized version of $\textbf{x}$ (i.e., $\widetilde{\mathbf{x}}$) is required even when this layer is frozen.  

\vspace{-0.1cm} 
The contribution of the last term in Eq. (\ref{LNI}) (i.e., $\sum_H (\textbf{g} * \widetilde{\mathbf{x}})$) is significant for large values of $\widetilde{\mathbf{x}}$ only. Therefore, the small values of $\widetilde{\mathbf{x}}$ can be discarded. Ideally, we want to have all the activations of this layer to be discarded when this layer is frozen. However, this will results in an accuracy degradation. As such, we prune away the small values in $\widetilde{\mathbf{x}}$ and keep the top $10\%$ largest values. In this way, the memory load of activations is significantly reduced. Of course, when this layer is not frozen, the backpropagation is performed without any approximation. Such a trick converts LayerNorm from a static layer to a semi-static one. It is worth mentioning that the indices to pruned activations are also stored along with activations. The details of the pruning procedure is provided in Appendix \ref{App_pruning}.

\vspace{-0.45cm}
\section{{\sc \textbf{SlimFit}}} \label{SF_sec}
\vspace{-0.25cm}
{\sc SlimFit} is a performance tool that exploits our ILS algorithm along with quantization and pruning to reduce the memory footprint of activations through an iterative freezing process. The total on-device GPU memory reduction of {\sc SlimFit} is a result of the memory reduction in both dynamic and static activations. Static activations contribute a fixed amount of memory whereas the memory usage of dynamic activations depends on the freezing rate. Given a high freezing rate, the memory footprint of activations and accordingly the total on-device GPU memory usage can be significantly reduced. The choice of freezing rate depends on the memory budget of the user. By increasing the freezing rate up to a certain point, there will be no performance degradation. However, using an extremely high freezing rate trades off memory for accuracy. Finding the breaking point of the method is task dependent and varies from one dataset to another.

\vspace{-0.35cm}
\section{Experimental Results}\label{ER_sec}
\vspace{-0.25cm}
We use the base version of BERT and ViT for our experiments. We fine-tune these two models using {\sc SlimFit} which is implemented on PyTorch. We evaluate BERT \cite{BERT} using the GLUE benchmark \cite{GLUE} and SQuAD 2.0 \cite{squad}. For ViT \cite{ViT}, we use CIFAR-10, CIFAR-100 and ImageNet datasets \cite{CIFAR, imagenet} for evaluation purposes. We discuss the memory usage of activations and the overall on-device GPU memory on the 32GB NVIDIA V100 GPU. We report the total on-device GPU memory usage using ``nvidia-smi''. For all the experiments in this section, we use $3$ epochs for fine-tuning. The details about the CV/NLP tasks, measurements and hyper-parameter settings are provided in Appendix \ref{AppHPS}.

\vspace{-0.25cm}
\subsection{Accuracy Evaluation on GLUE and SQuAD 2.0}
\vspace{-0.25cm}
To evaluate the language understanding ability of BERT models, the GLUE benchmark is formed by a series of downstream tasks including sentiment classification (SST-2), natural language inference (RTE, QNLI, and MNLI), paraphrase detection (MRPC, QQP, and STS-B), and linguistic acceptability (CoLA). We use Spearman correlation for STS-B, Matthew's correlation for CoLA, percentage accuracy for RTE, MRPC, SST-2, QQP, QNLI and MNLI$_{m }$, and F1 score for SQuAD 2.0. In this work, we fine-tune the BERT-base model using {\sc SlimFit} on the downstream tasks of the GLUE benchmark as well as the question answering task on SQuAD 2.0. Table \ref{tab3} shows the accuracy on the validation set of the aforementioned tasks and memory usage of {\sc SlimFit} compared to the baseline. The results of the baseline were obtained without freezing. We report the results associated with the highest freezing rate that can achieve a similar accuracy to that of the baseline by varying the learning rate. The experimental results on the GLUE benchmark show that up to $95\%$ of dynamic activations can be discarded with up to $0.4\%$ accuracy degradation, leading to an average of 1.9GB reduction in the total on-device GPU memory usage. On the other hand, while fine-tuning SQuAD 2.0 without freezing requires the minimum of 2 32GB NVIDIA V100 GPUs on a batch size of 128, {\sc SlimFit} enables its fine-tuning on a single 32GB NVIDIA V100 GPU, reducing the total on-device memory requirement of such a task from 58.5GB down to 19.1GB ($3.1\times$ reduction).

\begin{table}
\center
\caption{The accuracy and memory performance of {\sc SlimFit} on the GLUE benchmark and SQuAD 2.0. The batch size of $32$ and $128$ were used for GLUE benchmark and SQuAD 2.0, respectively.}
\scalebox{0.67}{
\input{tab3.tex}}
\label{tab3}
\vspace{-0.4cm}
\end{table}

\begin{figure}[t]
\vspace{-0.4cm}
\center
\scalebox{0.75}{
\input{fig7.tex}}
\vspace{-0.2cm}
\caption{The total on-device GPU memory usage of {\sc SlimFit} compared to the baseline across different batch sizes including 32, 64 and 128 on NLP and CV datasets.}
    \label{fig10}
    \vspace{-0.5cm}
\end{figure}
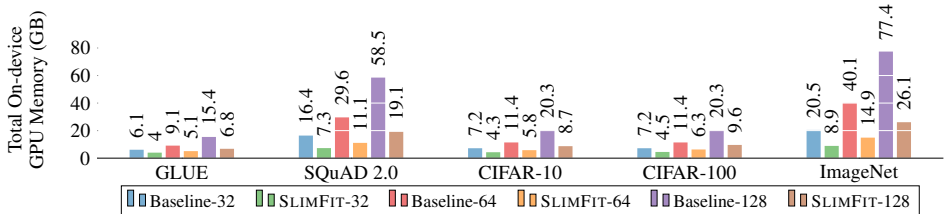

Figure \ref{fig10} shows the total on-device GPU memory usage of BERT when fine-tuned using {\sc SlimFit} for different batch sizes at the freezing rate of $95\%$ on the GLUE benchmark and $80\%$ on SQuAD 2.0. According to the experimental results, {\sc SlimFit} enables a reduction ranging from $1.5\times$ to $3.1\times$ in the total on-device GPU memory on NLP tasks. The reduction in the total on-device memory usage is more significant for larger batch sizes since the activations dominate the memory footprint.

\vspace{-0.35cm}
\subsection{Accuracy Evaluation on CIFAR and ImageNet}
\vspace{-0.25cm}
To assess the effectiveness of our method on CV tasks, we fine-tune the ViT-base model on CIFAR-10, CIFAR-100 and ImageNet datasets. We use the test set of CIFAR-10/CIFAR-100 and the validation set of ImageNet to evaluate their accuracy on ViT. Table \ref{tab4} shows that {\sc SlimFit} can fine-tune the ViT-base model with the freezing rate of up to $95\%$ with up to $0.3\%$ loss in accuracy while significantly reducing the overall on-device GPU memory usage. More specifically, {\sc SlimFit} reduces the overall memory usage of the fine-tuning process on CIFAR-10 from 7.2GB to 4.3GB ($1.7\times$ reduction) with a batch size of 32, on CIFAR-100 from 7.2GB to 4.5GB ($1.6\times$ reduction) with a batch size of 32, and on ImageNet from 77.4GB to 26.1GB ($3\times$ reduction) with a batch size of 128. Fig. \ref{fig10} also shows the total on-device GPU memory usage of {\sc SlimFit} across different batch sizes on CV tasks.

\begin{table}[t]

\caption{The top-1 accuracy and memory performance of {\sc SlimFit} on CV benchmarks using a batch size of $32$ for CIFAR datasets and $128$ for ImageNet dataset.}
\vspace{-0.2cm}
\center
\scalebox{0.7}{
\input{tab4_1.tex}}
\label{tab4}
\vspace{-0.3cm}
\end{table}

\vspace{-0.35cm}
\section{Ablation Studies}
\vspace{-0.25cm}
In this section, we study different aspects of {\sc SlimFit} in fine-tuning of transformer-based models through a series of ablation studies. Due to limited space, we discuss the impact of quantization/pruning and total wall-clock time in Appendix \ref{quant_pruning} and Appendix \ref{WCT}, respectively. For all the experiments in this section, we use a batch size of 32 and $3$ epochs for fine-tuning.

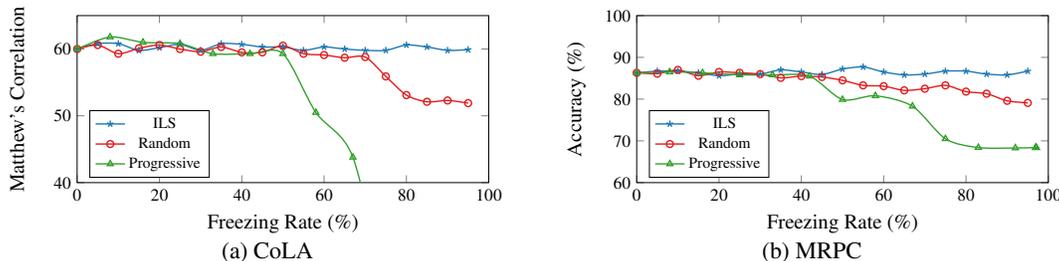
\begin{figure}[b]
\vspace{-0.35cm}
\center
\begin{subfigure}[t]{0.5\textwidth}
\scalebox{0.65}{
\input{fig6a.tex}}
\vspace{-0.2cm}
\caption{CoLA}
\label{fig6a}
    \end{subfigure}%
\hfill
    \begin{subfigure}[t]{0.5\textwidth}
        \center
        \scalebox{0.65}{
        \input{fig6b.tex}}

        \vspace{-0.2cm}
\caption{MRPC}
\label{fig6b}
    \end{subfigure}%
    \vspace{-0.25cm}
\caption{The trade-off curve between accuracy and freezing rate for three different iterative freezing approaches (i.e., ILS, random and progressive methods) on (a) CoLA  and (b) MRPC datasets.}
\label{fig6}
\vspace{-0.35cm}
\end{figure}

\vspace{-0.25cm}
\subsection{Accuracy vs Freezing Rate}
\vspace{-0.25cm}
In Section (\ref{ILSM}), we discussed that our ILS algorithm orchestrates the freezing schedule based on a simple rule: layers with largest distance values are updated whereas those with lowest distance values are kept frozen for the given freezing rate. Of course, such an iterative freezing approach trades off between accuracy and freezing rate. To better show this trade-off, we measured and illustrated accuracy of CoLA and MRPC datasets across different freezing rates in Fig. \ref{fig6}. 
The trade-off curve shows our ILS algorithm can maintain the accuracy at the same level of the baseline by freezing up to $95\%$ of layers.

Besides our ILS algorithm, the freezing schedule can be decided using random or progressive freezing approaches. In the random scheduling method, frozen layers are randomly selected at each iteration. In the progressive approach, on the other hand, early layers are progressively kept frozen whereas later layers are being updated throughout the fine-tuning process. Among these approaches, our ILS algorithm significantly stands out in terms of both accuracy and freezing rate as shown in Fig. \ref{fig6}. The
\begin{wrapfigure}{r}{0.68\textwidth}
\vspace{-0.32cm}
\center
\begin{subfigure}[t]{0.35\textwidth}
\includegraphics[clip, scale=0.4, trim=2cm 0cm 3cm 1cm]{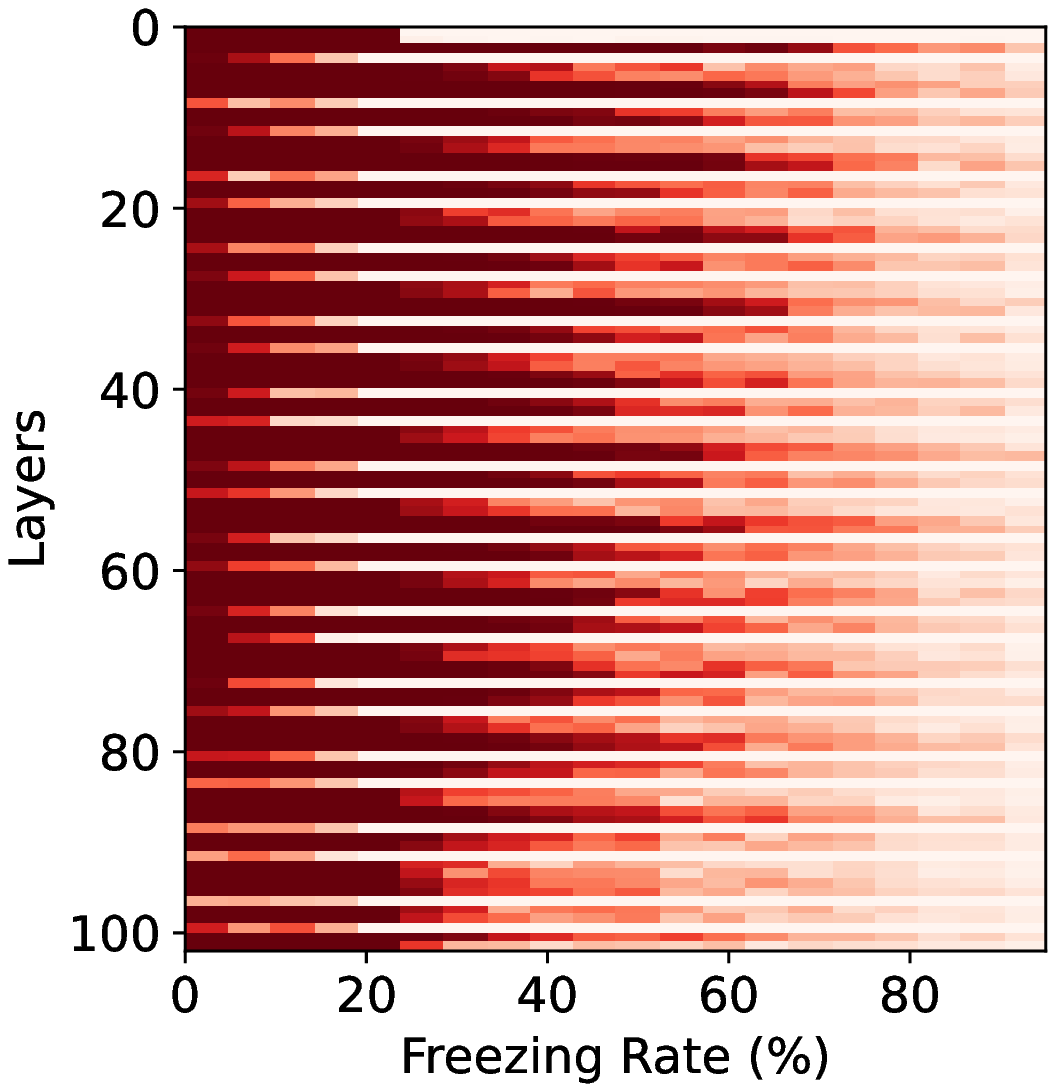}
\vspace{-0.15cm}
\caption{CoLA}
\label{fig7a}
    \end{subfigure}%
    \begin{subfigure}[t]{0.35\textwidth}
        \includegraphics[clip, scale=0.4, trim=2cm 0cm 3cm 1cm]{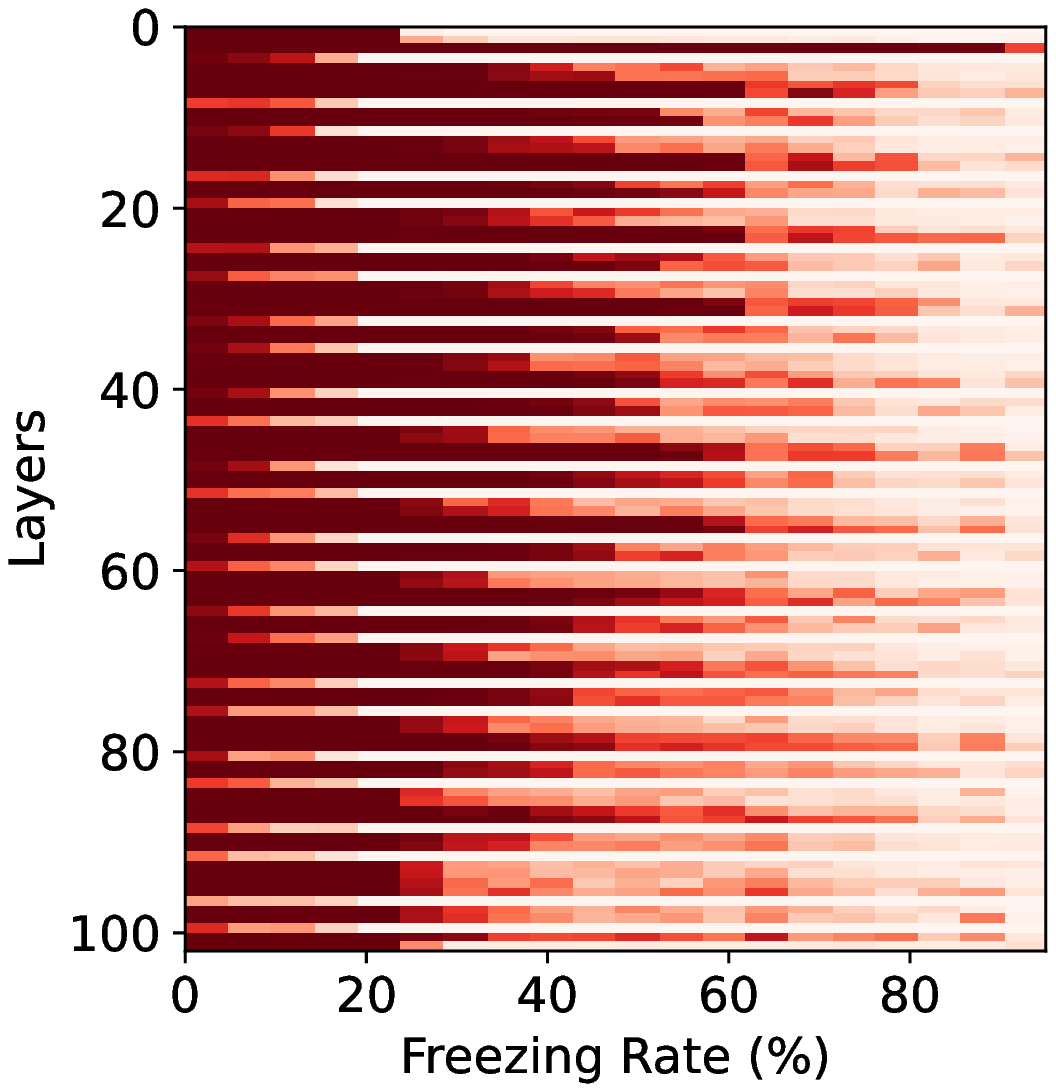}
\vspace{-0.15cm}
\caption{MRPC}
\label{fig7b}
    \end{subfigure}%
    \vspace{-0.25cm}
\caption{The frequency of update occurrence for each layer as a heatmap on (a) CoLA  and (b) MRPC datasets. The description of layers corresponding the indices are provided in Appendix \ref{AppHM}.}
\label{fig7}
\vspace{-0.25cm}
\end{wrapfigure}
reason behind its superior performance is that ILS allows more updates for layers with large distance values by keeping layers with minimal distance values frozen for a specific number of iterations. On the other hand, in the random approach, the layers are randomly selected to be updated. Therefore, layers with large distance values receive less number of updates in the random approach compared to ILS. Of course, the chance of layers with large distance values being randomly selected as active layers decreases as the freezing rate increases, which explains the accuracy gap between ILS and the random approach with freezing rate higher than 70\% freezing rate. In the progressive freezing approach, the early layers receive no update during the fine-tuning process, resulting in a significant accuracy degradation for large freezing rates.

\vspace{-0.25cm}
\subsection{Frequency of Update Occurrence}
\vspace{-0.25cm}
To visualize the frequency of update occurrence for each layer, we use a heatmap as shown in Fig. \ref{fig7} for both CoLA and MRPC datasets where larger counts are associated with darker colorings. As shown in the heatmap, the dense layers inside the MHA module receive more updates than other layers for both datasets. Moreover, the update patterns of these datasets are similar for small freezing rates whereas they become more task-specific for high freezing rates. In fact, the ILS algorithm prioritizes the update of some specific layers over others for high freezing rates.

\vspace{-0.25cm}
\section{Comparison With State-of-the-Art Techniques and Limitations}
\vspace{-0.25cm}
Next, we compare {\sc SlimFit} with state-of-the-art compression methods targeting memory reduction, i.e., GACT \cite{GACT} and DropIT \cite{DropIT}. Table \ref{tab6} summarizes the comparison results in terms of accuracy, memory and latency. For fair comparison, we measure their performance under the same framework and hyper-parameters (i.e., the batch size and the number of training epochs) during fine-tuning of BERT on CoLA. The experimental results of GACT and DropIT were obtained using their official PyTorch libraries. According to the experimental results, GACT shows the lowest memory amount for activations. However, in terms of on-device GPU memory usage, {\sc SlimFit} outperforms GACT. In terms of accuracy, all models show a comparable accuracy on CoLA w.r.t. the baseline. Finally, in terms of speed, {\sc SlimFit} shows the fastest fine-tuning speed among existing works while it still falls short w.r.t. the baseline (see Appendix \ref{WCT} for more details on {\sc SlimFit}'s computing speed). Despite the better accuracy of {\sc SlimFit} on CoLA, it shows up to $0.4\%$ degradation in accuracy across different CV/NLP tasks which is another limitation of {\sc SlimFit} besides its fine-tuning speed w.r.t. the baseline.

\begin{table}
\center
\caption{Comparison with state-of-the-arts when fine-tuning BERT on CoLA dataset.}
\scalebox{0.7}{
\input{tab6_1.tex}}
\label{tab6}
\vspace{-0.6cm}
\end{table}

\vspace{-0.3cm}
\section{Conclusion}
\vspace{-0.25cm}
In this paper, we presented a performance tool called {\sc SlimFit} that reduces the memory usage of activations and accordingly the overall on-device GPU memory usage of transformer-based models through an iterative freezing of layers during fine-tuning. {\sc SlimFit} adopts an inter-layer scheduling method to orchestrate the freezing schedule at each iteration. To balance the number of activations across all layers and to reduce the memory usage of static activations, {\sc SlimFit} uses quantization and pruning for a few specific layers. We evaluated the performance of {\sc SlimFit} across different NLP and CV tasks. We showed that {\sc SlimFit} significantly reduces the on-device GPU memory usage of the fine-tuning process by up to $3.1\times$ when using a batch size of $128$.

\bibliography{example_paper}
\bibliographystyle{IEEEtran}

\newpage
\appendix
\onecolumn
\section*{Appendix}

\section{Memory Management}\label{mem_management}
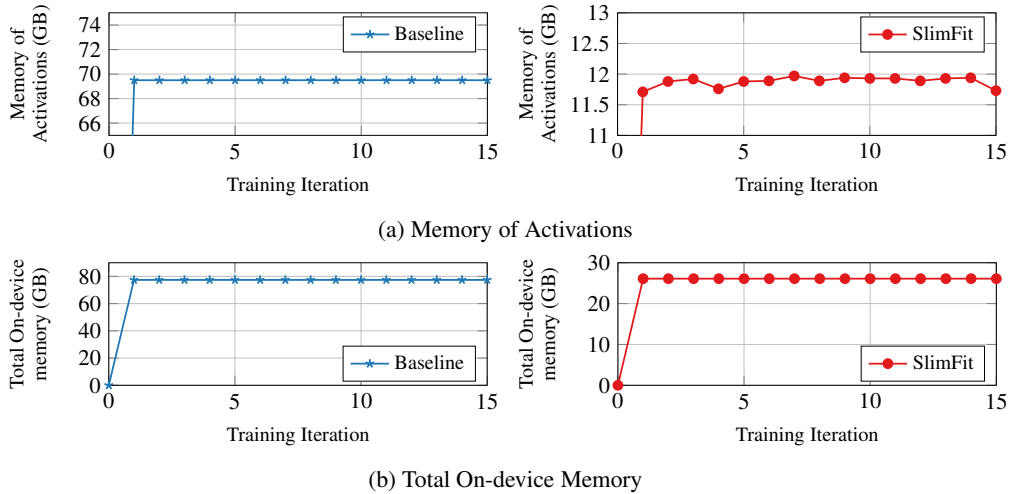
\begin{figure}[h]
\center
\begin{subfigure}[t]{\textwidth}
\center
\scalebox{0.85}{
\input{fig9a.tex}}
\caption{Memory of Activations}
\label{fig9a}
    \end{subfigure}%
\vfill
    \begin{subfigure}[t]{\textwidth}
        \center
        \scalebox{0.85}{
        \input{fig9b.tex}}
\caption{Total On-device Memory}
\label{fig9b}
    \end{subfigure}%
\caption{
The total on-device GPU memory and memory of activations during different training iterations when fine-tuning  ViT on ImageNet with a batch size of 128 with the freezing rate of $95\%$ compared to the baseline. {\sc SlimFit} balances the memory usage of activations using freezing to reduce the total on-device memory usage of the fine-tuning process. While the memory usage of activations changes at each iteration when using {\sc SlimFit}, the changes are relatively small thanks to the load-balancing technique described in Section \ref{ILLB_sec}.}
\label{fig9}
\end{figure}

The on-device memory of modern GPUs is limited to a few tens of gigabytes depending on their model (e.g., 32GB NVIDIA V100). If the memory requirement of the training/fine-tuning of neural networks goes beyond the available memory on GPUs, an out-of-memory error will occur. The memory requirement of the under-run code on GPUs can be viewed by ``nvidia-smi''. For a training/fine-tuning process, this memory requirement is determined by the size of the model, cached activations, gradients, gradient moments from the optimizer and CUDA contents after the first training iterations. It is worth mentioning that the memory usage of the training/fine-tuning process remains constant after the first iteration if the following iterations are performing the same computations (see Fig. \ref{fig9}). If the memory requirement of each iteration is different from others, PyTorch reports the memory requirement of the iteration using the maximum memory among other iterations as the total on-device GPU memory usage of the program in ``nvidia-smi'' \cite{pytorch_memory}. Therefore, the unused memory of tensors in progressive memory optimization over training iterations will still show as used in ``nvidia-smi''. To reduce the overall memory usage of the training/fine-tuning process given the above explanations, we need to balance the memory usage of all iterations from the first iterations to the last one (see Fig. \ref{fig9}). {\sc SlimFit} aims at reducing the overall memory usage of large transformer-based models during fine-tuning.

\section{Comparison with Existing Freezing Approaches} \label{Freezing_comp}
Here, we describe the main differences between {\sc SlimFit} and other freezing approaches including SmartFRZ \cite{smartfrz}, PipeTransformer \cite{pipetrans} and AutoFreeze \cite{AutoFreeze}. The main difference is that the aforementioned works mainly focus on exploiting freezing to accelerate the training/fine-tuning process. Conceptually, SmartFRZ, PipeTransformer and AutoFreeze progressively freeze layers as the training process proceeds. In these methods, the first training iteration starts without freezing where all layers are updated. In the following iterations, these methods then progressively start freezing from the early layers down to the latest layers in the model in an orderly fashion. For instance, AutoFreeze performs the first epoch without freezing, the second epoch while freezing the first 5 layers, the third epoch while freezing the first 8 layers and the fourth epoch while freezing the first 11 layers when fine-tuning BERT. In this example, the memory and computation requirement of each epoch is different from others as each epoch presents a different degree of freezing. This allows to exploit the unused computing and memory resources to further accelerate the process by increasing batch sizes as the memory decreases throughout the training iterations \cite{AutoFreeze} or increasing data-parallel width through pipelining \cite{pipetrans}. Since the first training iteration (or even epoch) of these methods performs the training process without freezing, their overall memory requirement reported by ``nvidia-smi'' is similar to that of training without freezing as discussed in Appendix \ref{mem_management}. In other words, the under-use GPU must still be able to meet the memory requirement of training without freezing. For instance, fine-tuning of ViT with a batch size of 128 on a single 32GB NVIDIA V100 using such methods results in an out-of-memory error.

{\sc SlimFit}, on the other hand, focuses on reducing the overall memory requirement of the fine-tuning process using freezing. As opposed to the aforementioned methods (i.e., SmartFRZ, PipeTransformer and AutoFreeze), {\sc SlimFit} freezes layers at every single training iterations from the first iteration to the last one with a fixed freezing rate. With load-balancing using quantization, {\sc SlimFit} ensures that the memory requirement of every single iteration remains roughly the same throughout the fine-tuning process. This enables {\sc SlimFit} performing memory-intensive fine-tuning processes with large batch sizes on a single 32GB GPU such as ViT on ImageNet with a batch size of 128  while this normally requires three 32GB GPUs.

\section{Architecture of Transformer-based Models} \label{Arch}
Fig. \ref{fig12} shows the overall architecture of transformer-based models including an initial embedding layer, followed by repeated blocks of multi-head attention (MHA) and feed-forward network (FFN). The details of each layer inside the MHA and FFN modules are provided in Table \ref{tab8}.

\begin{minipage}[!t]{\textwidth}
  \begin{minipage}[b]{0.4\textwidth}
    \centering
    \scalebox{0.8}{
\input{fig2.tex}}

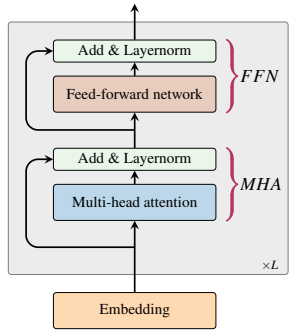
\captionof{figure}{The main architecture of BERT. Note that ViT has a similar architecture with LayerNorms located before the MHA block. $L$ denotes the number of attention layers.}
    \label{fig12}
  \end{minipage}
  \hfill
  \begin{minipage}[b]{0.57\textwidth}
    \centering
    \scalebox{0.8}{
\input{tab8.tex}}
\captionof{table}{The details of layers in MHA and FFN modules of BERT where $B$, $T$, $H$ denote the batch size, sequence length, hidden size, respectively. ViT has the same structure with different descriptions.}
    \label{tab8}
    \end{minipage}
  \end{minipage}

\section{Theoretical Analysis}
\subsection{Convergence Analysis} \label{conv_analysis}
In this section, we provide a convergence analysis for our freezing strategy. More precisely, we prove convergence of stochastic gradient descent (SGD) when considering freezing during update iterations. Given the loss function $f$, we assume that the parameters are initialized with some value and denoted as the vector $\textbf{w}_0 \in \mathbb{R}^d$. Given the training example, the parameters are updated by
\begin{equation}
    \textbf{w}_{t+1} = \textbf{w}_t - \gamma_t \nabla f(\textbf{w}_t),
    \label{update_eq}
\end{equation}
where $\textbf{w}_t$ denotes the parameter vector at time $t$, $\gamma_t$ is the learning rate, and $\nabla f$ represents the gradient of the loss function. We assume that the magnitude of the gradient samples are bounded by a constant $G > 0$ for all $\textbf{x}$ in the space such that
\begin{equation}
    ||\nabla f(\textbf{x})|| \leq G.
    \label{cond1}
\end{equation}
Also, we assume that there exists a constant $L > 0$ for any vector $\textbf{u} \in \mathbb{R}^d$ where we have
\begin{equation}
    |\textbf{u}^T \nabla^2 f(\textbf{x})\textbf{u}| \leq L ||\textbf{u}||^2.
    \label{cond2}
\end{equation}
Given Eq. (\ref{cond1}) and Eq. (\ref{cond2}), performing Taylor expansion on Eq. (\ref{update_eq}) similar to \cite{SGDproof} results in 
\begin{equation}
    \mathbb{E}[f(\textbf{w}_{t+1})] \leq \mathbb{E}[f(\textbf{w}_t)] - \gamma_t \mathbb{E}[||\nabla f(\textbf{w}_t)||^2] + \dfrac{\gamma_t^2G^2L}{2},
    \label{taylor_eq}
\end{equation}
where $\mathbb{E}$ denotes the expected value.

Now, let us assume that the layer containing the parameter vector is frozen at the training iteration $t$. In this case, $\nabla f(\textbf{w}_t)$ is equal to $0$ and consequently $\textbf{w}_{t+1}$ is equal to $\textbf{w}_t$. In this freezing scenario, Eq. (\ref{taylor_eq}) still holds true since $\dfrac{\gamma_t^2G^2L}{2}$ is greater than $0$.

By rearranging the terms in Eq. (\ref{taylor_eq}), summing over $T$ iterations and telescoping the sum, we obtain
\begin{align}
    \sum_{t=0}^{T-1} \gamma_t \mathbb{E}[||\nabla f(\textbf{w}_t)||^2] & \leq \sum_{t=0}^{T-1} (\mathbb{E}[f(\textbf{w}_t)] - \mathbb{E}[f(\textbf{w}_{t+1})]) + \sum_{t=0}^{T-1}\dfrac{\gamma_t^2G^2L}{2},\\
    & = f(\textbf{w}_0) - f(\textbf{w}_T) + \dfrac{G^2L}{2} \sum_{t=0}^{T-1} \gamma_t^2,\\
    & \leq f(\textbf{w}_0) - f(\textbf{w}_*) + \dfrac{G^2L}{2} \sum_{t=0}^{T-1} \gamma_t^2,
\end{align}
where $\textbf{w}_*$ indicates an optimal solution. Given the above inequality, we showed that the convergence proof of SGD remains intact while introducing freezing for specific training iterations.

\subsection{Backpropagation With a Frozen Layer} \label{backprop_frozen_layer}
Here, we provide a simple example demonstrating how gradients are backpropagated to the first layer of a neural network while its middle layer is frozen. To this end, let us perform the backpropagation using a 3-layer network as an example. Mathematically, the architecture of this network can be described as follows:
\begin{align}
    \textbf{y}_1 &= \textbf{x}\textbf{W}_1 + \textbf{b}_1,\\
    \textbf{y}_2 &= \textbf{y}_1\textbf{W}_2 + \textbf{b}_2,\\
    \textbf{y}_3 &= \textbf{y}_2\textbf{W}_3 + \textbf{b}_3,
\end{align}
where $\textbf{W}_1$, $\textbf{W}_2$, $\textbf{W}_3$, $\textbf{b}_1$, $\textbf{b}_2$ and $\textbf{b}_3$ are the weights and biases of the network. In this example, $\textbf{x}$, $\textbf{y}_1$ and $\textbf{y}_2$ are inputs to the first layer, the second layer and the third layer, respectively. Now, let us derive the backpropagation equations with the loss
$\mathcal{L}$ using the chain rule as follows (please note that we obtain $\dfrac{\partial \mathcal{L}}{\partial \textbf{y}_3}$ by computing the loss where $\partial$ denotes the partial derivative):
\begin{align}
    \dfrac{\partial \mathcal{L}}{\partial \textbf{W}_3} & = \dfrac{\partial \textbf{L}}{\partial \textbf{y}_3} \dfrac{\partial \textbf{y}_3}{\partial \textbf{W}_3} = \dfrac{\partial \mathcal{L}}{\partial \textbf{y}_3} \textbf{y}_2, \label{partial_1} \\ 
    \dfrac{\partial \mathcal{L}}{\partial \textbf{b}_3} & = \dfrac{\partial \mathcal{L}}{\partial \textbf{y}_3} \dfrac{\partial \textbf{y}_3}{\partial \textbf{b}_3} = \dfrac{\partial \mathcal{L}}{\partial \textbf{y}_3} 1 = \dfrac{\partial \mathcal{L}}{\partial \textbf{y}_3}, \\
    \dfrac{\partial \mathcal{L}}{\partial \textbf{W}_2} & = \dfrac{\partial \mathcal{L}}{\partial \textbf{y}_3} \dfrac{\partial \textbf{y}_3}{\partial \textbf{y}_2} \dfrac{\partial \textbf{y}_2}{\partial \textbf{W}_2} = \dfrac{\partial \mathcal{L}}{\partial \textbf{y}_3} \textbf{W}_3^T \textbf{y}_1, \label{partial_2} \\ 
    \dfrac{\partial \mathcal{L}}{\partial \textbf{b}_2} & = \dfrac{\partial \mathcal{L}}{\partial \textbf{y}_3}  \dfrac{\partial \textbf{y}_3}{\partial \textbf{y}_2} \dfrac{\partial \textbf{y}_2}{\partial \textbf{b}_2} = \dfrac{\partial \mathcal{L}}{\partial \textbf{y}_3} \textbf{W}_3^T 1 = \dfrac{\partial \mathcal{L}}{\partial \textbf{y}_3} \textbf{W}_3^T, \\
    \dfrac{\partial \mathcal{L}}{\partial \textbf{W}_1} & = \dfrac{\partial \mathcal{L}}{\partial \textbf{y}_3} \dfrac{\partial \textbf{y}_3}{\partial \textbf{y}_2} \dfrac{\partial \textbf{y}_2}{\partial \textbf{y}_1} \dfrac{\partial \textbf{y}_1}{\partial \textbf{W}_1} = \dfrac{\partial \mathcal{L}}{\partial \textbf{y}_3} \textbf{W}_3^T \textbf{W}_2^T \textbf{x}, \label{partial_3} \\ 
    \dfrac{\partial \mathcal{L}}{\partial \textbf{b}_1} & = \dfrac{\partial \mathcal{L}}{\partial \textbf{y}_3} \dfrac{\partial \textbf{y}_3}{\partial \textbf{y}_2} \dfrac{\partial \textbf{y}_2}{\partial \textbf{y}_1} \dfrac{\partial \textbf{y}_1}{\partial \textbf{b}_1} = \dfrac{\partial \mathcal{L}}{\partial \textbf{y}_3} \textbf{W}_3^T \textbf{W}_2^T 1 = \dfrac{\partial \mathcal{L}}{\partial \textbf{y}_3} \textbf{W}_3^T \textbf{W}_2^T.\label{partial_4}
\end{align}
Given the above equations, to update the network weights (i.e., $\textbf{W}_1$, $\textbf{W}_2$, and $\textbf{W}_3$), we need to store $\textbf{x}$, $\textbf{y}_1$ and $\textbf{y}_2$ during the forward computations since they are required in Eq. (\ref{partial_1}), Eq. (\ref{partial_2}) and Eq. (\ref{partial_3}) during the back computations. 

Now, suppose the middle layer is frozen. In this case, there is no need to compute Eq. (\ref{partial_2}) and therefore there is no need to store $\textbf{y}_1$ during the forward computations. Of course, discarding $\textbf{y}_1$ does not affect the backward computations of the first layer since Eq. (\ref{partial_3}) and Eq. (\ref{partial_4}) are independent of $\textbf{y}_1$.



\section{Conversion Between 8-bit Integer and 32-bit Floating-point} \label{App_conv}
Algorithm \ref{alg2} shows the conversion process between 8-bit fixed-point and 32-bit floating-point formats. It is worth mentioning that the same procedure can be used for the conversion between 4-bit fixed-point and 32-bit floating-point formats. Moreover, the quantization function is used to compress the cached tensors during the forward propagation only. Of course, both the forward and backward computations are still performed using 32-bit floating-point computations as shown in Algorithm \ref{alg4} where the ``compress'' function in this case is the quantization function (i.e., the conversion from 32-bit floating-point to 8-bit integer) and the ``decompress'' function performs the reverse computations (i.e., the conversion from 8-bit integer to 32-bit floating-point).

\begin{algorithm}[h]
   \caption{The conversion between 8-bit integer and 32-bit floating-point.}
   \label{alg2}
\input{alg2.tex}
\end{algorithm}

\begin{algorithm}[b]
   \caption{The description of the pruning process during the forward computations and the restoring process during the backward computations.}
   \label{alg5}
\input{alg5.tex}
\end{algorithm}

\vspace{-0.5cm}
\section{Pruning Algorithm} \label{App_pruning}
The pruning algorithm is performed in a few steps. In the first step, the input vector is sorted from largest to smallest values along with their indices and the size of the dense vector. We then only keep and cache the top $10\%$ largest values of the input vector for the backward computations as the second step. It is worth mentioning that pruning beyond $90\%$ results in a significant accuracy degradation. During backpropagation, we create a zero-valued tensor using the size of the dense vector and then replace zero values with the top $10\%$ largest values using their corresponding indices. Algorithm \ref{alg5} shows the pruning process during the forward computations and the restoring process during the backward computations. It is worth mentioning that the pruning function is used to compress the cached tensors during the forward propagation only. Of course, both the forward and backward computations are still performed using 32-bit floating-point computations as shown in Algorithm \ref{alg4} where the ``compress'' function in this case is the pruning function and the ``decompress'' function performs the restoring computations.

\section{Details of CV/NLP Tasks, Measurements and Hyper-parameter Settings} \label{AppHPS}
For language understanding tasks and CV tasks, we used BERT-base-cased and ViT-base throughout this paper, respectively. The BERT-base and ViT-base pre-trained on ImageNet-21k were configured according to \cite{BERT} and \cite{ViT}, respectively. We used AdamW ($\beta_1 = 0.9$, $\beta_2 = 0.999$ and $L2$ weight decay of 0.01) as the optimizer and linear decay of the learning rate with warmup ranging from 0 to 0.1 for both models. We evaluate BERT-base on several downstream tasks from the GLUE benchmark and SQuAD 2.0. We use Spearman correlation for STS-B, Matthews correlation for CoLA, accuracy for RTE, MRPC, SST-2 QQP, QNLI and MNLI$_m$ (matched), and F1 score for SQuAD 2.0. For the downstream tasks from the GLUE benchmark, we used the sequence length of 128 whereas we adopted the sequence length of 384 for the question answering task on SQuAD 2.0. For CIFAR-10, CIFAR-100 and ImageNet, we use top-1 accuracy as our evaluation metric. For the image classification tasks, we used the patch size of 16 with the resolution of 224 for CIFAR-10/CIFAR-100 and the resolution of 384 for ImageNet. Depending on the task, the learning rate varies from 4e-5 to 1.8e-4. For all the experiments in this paper, we used 3 epochs for fine-tuning. The hyper-parameter settings of each task are summarized in Table \ref{tab7}. It is worth mentioning that ViT models can also be fine-tuned using SGD. However, fine-tuning ViT models using SGD requires more epochs w.r.t. AdamW to obtain a similar accuracy.

In this paper, we measured our experimental results directly from 32GB NVIDIA V100 GPU(s) without any memory swapping between CPU and GPU(s). The total on-device GPU memory usage of the fine-tuning process is measured using ``nvidia-smi''. We measured the wall-clock time (i.e., latency) of the fine-tuning process using the CUDA's event API in PyTorch (i.e., ``torch.cuda.Event''). The memory footprint of activations on the GPU(s) was measured using the PyTorch's memory management API (i.e., ``torch.cuda.memory\_allocated'').

\begin{table}[h]
\center
\caption{The hyper-parameter settings of each NLP/CV task.}
\scalebox{0.875}{
\input{tab7.tex}}
\label{tab7}
\end{table}

\section{Impact of Quantization and Pruning} \label{quant_pruning}
In this work, we used quantization and pruning for a few specific layers to balance the number of activations across all layers and to reduce the memory footprint of static activations. We used 8-bit quantization for the activations of the imbalanced linear layer and MatMul. We also quantized the activations of GELU using 4 bits. The pruning of LayerNorm was performed when this layer is kept frozen. It is worth mentioning that both quantization and pruning have no impact on the forward computations. They are only used to compress activations for caching. To show the impact of such compression methods, we report the accuracy evaluation of BERT on CoLA and MRPC datasets with and without quantization or pruning in Table \ref{tab5}. The experimental results show no notable performance loss due to the compression techniques. 

\begin{table}[h]
\center
\caption{The impact of quantization and pruning on the accuracy evaluation.}
\scalebox{0.9}{
\input{tab5.tex}}
\label{tab5}
\vspace{-0.5cm}
\end{table}

\section{Discussion on Wall-Clock Time} \label{WCT}
Compared to training without freezing, {\sc SlimFit} introduces extra computations and also skips weight gradient computations for the frozen layers at the same time. The main source of computational overhead in {\sc SlimFit} is quantization and pruning of activations. The quantization overhead is due to the conversion between different precision levels (i.e., between 8 bits and 32-bit floating-point format) as discussed in Appendix \ref{App_conv}. Pruning also requires sorting of values to keep their top $10\%$ largest values, which causes an additional computational overhead. Computing the weight distance metric is another source of computational overhead. 

\begin{algorithm}[h]
   \caption{The description of skipping weight gradient computations when the layer is frozen. In this example, we assume the activations of the frozen layer require compression (e.g., an imbalanced linear layer or LayerNorm). Activations are denoted as ``input'' and are cached using either quantization or pruning depending on the type of the layer as a compression method. The compression and decompression functions are denoted as ``compress'' and ``decompress''. Since weights are defined as ``Parameter'' in PyTorch, caching weights does not introduce any extra memory. }
   \label{alg4}
\input{alg4.tex}

\end{algorithm}

\begin{figure}[b]
\center
\scalebox{0.8}{
\input{fig8.tex}}
\caption{Wall-clock time of fine-tuning ViT on ImageNet with a batch size of 32 across different freezing rates.}
\label{fig8}
\end{figure}
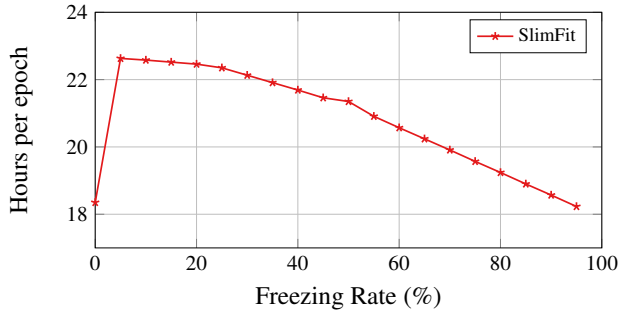

On the other hand, {\sc SlimFit} skips the weight gradients computations of frozen layers using PyTorch ``requires\_grad'' as shown in Algorithm \ref{alg4}. When an activate layer is frozen, there is no need to compute its weight gradients as discussed in Appendix \ref{backprop_frozen_layer}, which reduces the wall-clock time. The amount of speedup due to the skipped computations highly depends on the hyper-parameters of the networks such as freezing rate. Therefore, the wall-clock time of each network varies from one to another depending on the hyper-parameters. For instance, Fig. \ref{fig8} shows the wall-clock time of fine-tuning ViT on ImageNet using a batch size of 32 across different freezing rates. According to the experimental results, the computational overhead of {\sc SlimFit} is dominant for small freezing rates. However, as the freezing rate increases, the speedup of the skipped gradient computations overcomes the computational overhead of {\sc SlimFit} where {\sc SlimFit} with the freezing rate of $95\%$ results in a similar wall-clock time as of the baseline. It is worth mentioning that the baseline is the point at the freezing rate of $0$ where no freezing was used during the fine-tuning process.

\section{Description of Layers in the Heatmap} \label{AppHM}
The description of layers associated to the indices in Fig. \ref{fig7} is provided in Algorithm \ref{alg3}. It is worth mentioning that the layers denoted by ``bert.encoder.layer[i].attention'' belong to the MHA module whereas the remaining layers inside the loop belong to the FFN module.

\begin{algorithm}[h]
   \caption{The description of layers associated to the indices in Fig. \ref{fig7}.}
   \label{alg3}
\input{alg3.tex}
\end{algorithm}

\end{document}

%% file: fig1.tex
\small

\begin{tikzpicture}
  \centering
  \begin{axis}[
        ybar, axis on top,
        height=4cm, width=7cm,
        bar width=0.25cm,
        ymajorgrids, tick align=inside,
        major grid style={draw =white},
        enlarge y limits={value=.1,upper},
        ymin=0, ymax=16,
        axis x line*=bottom,
        axis y line*=left,
        y axis line style={opacity=0},
        tickwidth=0pt,
        enlarge x limits={abs = 1.5cm},
        x = 2cm,
        legend style={font=\footnotesize,
            at={(0.5,-0.2)},
            anchor=north,
            legend columns=-1,
            /tikz/every even column/.append style={column sep=0.3cm}
        },
        ylabel style ={align=center,at={(axis description cs:0.075,0.5)}, font=\footnotesize},
        xlabel style ={font=\small},
        tick label style={font=\tiny\scriptsize},
        ylabel= Memory (GByte),
        symbolic x coords={
         Batch Size: 32, Batch Size: 64, Batch Size: 128},
       xtick=data,
       every node near coord/.append style={scale = 0.8},
       nodes near coords=\rotatebox{90}{
        \footnotesize\pgfmathprintnumber[precision=1]{\pgfplotspointmeta}
       }
    ]
    \addplot [draw=none, fill=Paired-1!60] coordinates {
      (Batch Size: 32, 6.1)
      (Batch Size: 64, 9.1)
      (Batch Size: 128, 15.4) };
   \addplot [draw=none,fill=Paired-3!60] coordinates {
      (Batch Size: 32, 3.2)
      (Batch Size: 64, 6.4)
      (Batch Size: 128, 12.8) };
   \addplot [draw=none,fill=Paired-5!60] coordinates {
      (Batch Size: 32, 0.4)
      (Batch Size: 64, 0.4)
      (Batch Size: 128, 0.4) };
    \addplot [draw=none,fill=Paired-7!60] coordinates {
      (Batch Size: 32, 2.4)
      (Batch Size: 64, 2.3)
      (Batch Size: 128, 2.2) };

    \legend{Total, Activations,
          Parameters, Others}
  \end{axis}
  \end{tikzpicture}

%% file: fig3a.tex
\begin{tikzpicture}
\begin{axis}[
    xlabel={Iteration},
    ymin=0,
    ymax=6,
    xmin = 0,
    xmax = 804,
    ylabel=Distance Value,
    legend pos=north east,
    label style={font=\large},
    legend style={font=\footnotesize},
    legend entries={Query Weights (layer \#$1$), Query Weights (layer \#$5$), Query Weights (layer \#$11$) },
    ]
    \addplot[color=Paired-1, thick, smooth,each nth point={7}] %
    table[x=i,
          y=x,
          col sep=space]
          {./cola.txt};
     \addplot[color=Paired-5, thick, smooth,each nth point={7}] %
    table[x=i,
          y=y,
          col sep=space]
          {./cola.txt};
     \addplot[color=Paired-3, thick, smooth,each nth point={7}] %
    table[x=i,
          y=z,
          col sep=space]
          {./cola.txt};
\end{axis}
\end{tikzpicture}

%% file: fig3b.tex
\begin{tikzpicture}
\begin{axis}[
    xlabel={Iteration},
    ymin=0,
    ymax=11,
    xmin = 0,
    xmax = 345,
    ylabel=Distance Value,
    legend pos=north east,
    label style={font=\large},
    legend style={font=\footnotesize},
    legend entries={ Query Weights (layer \#$1$), Query Weights (layer \#$5$), Query Weights (layer \#$11$) },
    ]
    \addplot[color=Paired-1, thick, smooth,each nth point={3}] %
    table[x=i,
          y=x,
          col sep=space]
          {./mrpc.txt};
     \addplot[color=Paired-5, thick, smooth,each nth point={3}] %
    table[x=i,
          y=y,
          col sep=space]
          {./mrpc.txt};
     \addplot[color=Paired-3, thick, smooth,each nth point={3}] %
    table[x=i,
          y=z,
          col sep=space]
          {./mrpc.txt};
\end{axis}
\end{tikzpicture}

%% file: fig2_1.tex
\begin{tikzpicture}[scale=.6, transform shape];

        \node [draw, fill=datemagenta!50, rotate=0, anchor=north west, minimum width=0.5cm, minimum height=1cm, rounded corners=.05cm,]  at ( 2,     0)    (rec1) {};
        \node [draw, fill=dateblue!50, rotate=0, anchor=north west, minimum width=0.5cm, minimum height=1cm, rounded corners=.05cm,]  at ( 2.5,     0)    (rec2) {};
        \node [draw, fill=datemagenta!50, rotate=0, anchor=north west, minimum width=0.5cm, minimum height=1cm, rounded corners=.05cm,]  at ( 3,     0)    (rec3) {};
        \node [draw, fill=dateblue!50, rotate=0, anchor=north west, minimum width=0.5cm, minimum height=1cm, rounded corners=.05cm,]  at ( 3.5,     0)    (rec4) {};
        \node [draw, fill=datemagenta!50, rotate=0, anchor=north west, minimum width=0.5cm, minimum height=1cm, rounded corners=.05cm,]  at ( 4,     0)    (rec5) {};
        \node [draw, fill=dateblue!50, rotate=0, anchor=north west, minimum width=0.5cm, minimum height=1cm, rounded corners=.05cm,]  at ( 4.5,     0)    (rec6) {};
        \node [draw, fill=dateblue!50, rotate=0, anchor=north west, minimum width=0.5cm, minimum height=1cm, rounded corners=.05cm,]  at ( 5,     0)    (rec7) {};
        \node [draw, fill=datemagenta!50, rotate=0, anchor=north west, minimum width=0.5cm, minimum height=1cm, rounded corners=.05cm,]  at ( 5.5,     0)    (rec8) {};

        \node [draw, fill=dateblue!50, rotate=0, anchor=north west, minimum width=0.5cm, minimum height=1cm, rounded corners=.05cm,]  at ( 8,     0)    (rec11) {};
        \node [draw, fill=datemagenta!50, rotate=0, anchor=north west, minimum width=0.5cm, minimum height=1cm, rounded corners=.05cm,]  at ( 8.5,     0)    (rec22) {};
        \node [draw, fill=dateblue!50, rotate=0, anchor=north west, minimum width=0.5cm, minimum height=1cm, rounded corners=.05cm,]  at ( 9,     0)    (rec33) {};
        \node [draw, fill=datemagenta!50, rotate=0, anchor=north west, minimum width=0.5cm, minimum height=1cm, rounded corners=.05cm,]  at ( 9.5,     0)    (rec44) {};
        \node [draw, fill=dateblue!50, rotate=0, anchor=north west, minimum width=0.5cm, minimum height=1cm, rounded corners=.05cm,]  at ( 10,     0)    (rec55) {};
        \node [draw, fill=datemagenta!50, rotate=0, anchor=north west, minimum width=0.5cm, minimum height=1cm, rounded corners=.05cm,]  at ( 10.5,     0)    (rec66) {};
        \node [draw, fill=datemagenta!50, rotate=0, anchor=north west, minimum width=0.5cm, minimum height=1cm, rounded corners=.05cm,]  at ( 11,     0)    (rec77) {};
        \node [draw, fill=dateblue!50, rotate=0, anchor=north west, minimum width=0.5cm, minimum height=1cm, rounded corners=.05cm,]  at ( 11.5,     0)    (rec88) {};

        \node [draw, fill=datemagenta!50, rotate=0, anchor=north west, minimum width=0.5cm, minimum height=1cm, rounded corners=.05cm,]  at ( 14,     0)    (rec111) {};
        \node [draw, fill=dateblue!50, rotate=0, anchor=north west, minimum width=0.5cm, minimum height=1cm, rounded corners=.05cm,]  at ( 14.5,     0)    (rec222) {};
        \node [draw, fill=dateblue!50, rotate=0, anchor=north west, minimum width=0.5cm, minimum height=1cm, rounded corners=.05cm,]  at ( 15,     0)    (rec333) {};
        \node [draw, fill=datemagenta!50, rotate=0, anchor=north west, minimum width=0.5cm, minimum height=1cm, rounded corners=.05cm,]  at ( 15.5,     0)    (rec444) {};
        \node [draw, fill=datemagenta!50, rotate=0, anchor=north west, minimum width=0.5cm, minimum height=1cm, rounded corners=.05cm,]  at ( 16,     0)    (rec555) {};
        \node [draw, fill=dateblue!50, rotate=0, anchor=north west, minimum width=0.5cm, minimum height=1cm, rounded corners=.05cm,]  at ( 16.5,     0)    (rec666) {};
        \node [draw, fill=datemagenta!50, rotate=0, anchor=north west, minimum width=0.5cm, minimum height=1cm, rounded corners=.05cm,]  at ( 17,     0)    (rec777) {};
        \node [draw, fill=dateblue!50, rotate=0, anchor=north west, minimum width=0.5cm, minimum height=1cm, rounded corners=.05cm,]  at ( 17.5,     0)    (rec888) {};

        \node [draw, fill=dateblue!50, rotate=0, anchor=north west, minimum width=0.5cm, minimum height=1cm, rounded corners=.05cm,]  at ( 22,     0)    (rec1111) {};
        \node [draw, fill=datemagenta!50, rotate=0, anchor=north west, minimum width=0.5cm, minimum height=1cm, rounded corners=.05cm,]  at ( 22.5,     0)    (rec2222) {};
        \node [draw, fill=datemagenta!50, rotate=0, anchor=north west, minimum width=0.5cm, minimum height=1cm, rounded corners=.05cm,]  at ( 23,     0)    (rec3333) {};
        \node [draw, fill=dateblue!50, rotate=0, anchor=north west, minimum width=0.5cm, minimum height=1cm, rounded corners=.05cm,]  at ( 23.5,     0)    (rec4444) {};
        \node [draw, fill=dateblue!50, rotate=0, anchor=north west, minimum width=0.5cm, minimum height=1cm, rounded corners=.05cm,]  at ( 24,     0)    (rec5555) {};
        \node [draw, fill=datemagenta!50, rotate=0, anchor=north west, minimum width=0.5cm, minimum height=1cm, rounded corners=.05cm,]  at ( 24.5,     0)    (rec6666) {};
        \node [draw, fill=dateblue!50, rotate=0, anchor=north west, minimum width=0.5cm, minimum height=1cm, rounded corners=.05cm,]  at ( 25,     0)    (rec7777) {};
        \node [draw, fill=datemagenta!50, rotate=0, anchor=north west, minimum width=0.5cm, minimum height=1cm, rounded corners=.05cm,]  at ( 25.5,     0)    (rec8888) {};

        \node [draw, fill=Paired-7!30, rotate=0, anchor=north west, minimum width=2cm, minimum height=1cm, rounded corners=.05cm,]  at ( 3,     2)    (slimfit1) {\large ILS};
        \node [draw, fill=Paired-7!30, rotate=0, anchor=north west, minimum width=2cm, minimum height=1cm, rounded corners=.05cm,]  at ( 9,     2)    (slimfit2) {\large ILS};
        \node [draw, fill=Paired-7!30, rotate=0, anchor=north west, minimum width=2cm, minimum height=1cm, rounded corners=.05cm,]  at ( 15,     2)    (slimfit3) {\large ILS};
        \node [draw, fill=Paired-7!30, rotate=0, anchor=north west, minimum width=2cm, minimum height=1cm, rounded corners=.05cm,]  at ( 23,     2)    (slimfit4) {\large ILS};

        \draw [->, >=stealth, thick, rounded corners=0.2cm, color = datemagenta] (rec8.east) -| ([xshift=-2.5cm] slimfit2.west) -- (slimfit2.west);
        \draw [->, >=stealth, thick, rounded corners=0.2cm, color = datemagenta] (rec88.east) -| ([xshift=-2.5cm] slimfit3.west) -- (slimfit3.west);
        \draw [->,  >=stealth, thick, rounded corners=0.2cm, color = datemagenta] (rec888.east) -| ([xshift=-4.5cm] slimfit4.west) -- (slimfit4.west);

        \node  at ( 4, -1.5)    (heads)  {\Large Transformer Model};
        \node  at ( 10, -1.5)    (heads)  {\Large Transformer Model};
        \node  at ( 16, -1.5)    (heads)  {\Large Transformer Model};
        \node  at ( 24, -1.5)    (heads)  {\Large Transformer Model};

        \node  at ( 4, 2.5)    (heads)  {\Large \textbf{Training Iteration} $\textbf{0}$};
        \node  at ( 10, 2.5)    (heads)  {\Large \textbf{Training Iteration} $\textbf{1}$};
        \node  at ( 16, 2.5)    (heads)  {\Large \textbf{Training Iteration} $\textbf{2}$};
        \node  at ( 24, 2.5)    (heads)  {\Large \textbf{Training Iteration} $\textbf{n-1}$};

        \draw [->, >=stealth, thick, rounded corners=0.2cm] (slimfit1.south) -- (rec2.north);
        \draw [->, >=stealth, thick, rounded corners=0.2cm] (slimfit1.south) -- (rec4.north);
        \draw [->, >=stealth, thick, rounded corners=0.2cm] (slimfit1.south) -- (rec6.north);
        \draw [->, >=stealth, thick, rounded corners=0.2cm] (slimfit1.south) -- (rec7.north);

        \draw [->, >=stealth, thick, rounded corners=0.2cm] (slimfit2.south) -- (rec11.north);
        \draw [->, >=stealth, thick, rounded corners=0.2cm] (slimfit2.south) -- (rec33.north);
        \draw [->, >=stealth, thick, rounded corners=0.2cm] (slimfit2.south) -- (rec55.north);
        \draw [->, >=stealth, thick, rounded corners=0.2cm] (slimfit2.south) -- (rec88.north);

        \draw [->, >=stealth, thick, rounded corners=0.2cm] (slimfit3.south) -- (rec222.north);
        \draw [->, >=stealth, thick, rounded corners=0.2cm] (slimfit3.south) -- (rec333.north);
        \draw [->, >=stealth, thick, rounded corners=0.2cm] (slimfit3.south) -- (rec666.north);
        \draw [->, >=stealth, thick, rounded corners=0.2cm] (slimfit3.south) -- (rec888.north);

        \draw [->, >=stealth, thick, rounded corners=0.2cm] (slimfit4.south) -- (rec1111.north);
        \draw [->, >=stealth, thick, rounded corners=0.2cm] (slimfit4.south) -- (rec4444.north);
        \draw [->, >=stealth, thick, rounded corners=0.2cm] (slimfit4.south) -- (rec5555.north);
        \draw [->, >=stealth, thick, rounded corners=0.2cm] (slimfit4.south) -- (rec7777.north);

        \draw [dashed, >=stealth, ultra thick, rounded corners=0.2cm, color= gray!75] (7, -2) -- (7,2.5);
        \draw [dashed, >=stealth, ultra thick, rounded corners=0.2cm, color= gray!75] (13, -2) -- (13,2.5);
        \draw [dashed, >=stealth, ultra thick, rounded corners=0.2cm, color= gray!75] (20, -2) -- (20,2.5);

\node [draw,fill=dateblue!50,rotate=0,anchor=north west, minimum width=0.5cm,minimum height=1cm, rounded corners=.05cm]  at ( 6, -2.5)    (l1){};	
\node  at ([xshift=1cm] l1.east)    ()  {\LARGE Frozen};

\node [draw,fill=datemagenta!50,rotate=0,anchor=north west, minimum width=0.5cm,minimum height=1cm, rounded corners=.05cm]  at ( 10, -2.5)    (l2){};	
\node  at ([xshift=1cm] l2.east)    ()  {\LARGE Active};

\draw [->, >=stealth, thick, rounded corners=0.2cm]  ( 14, -3) -- (15,-3) ;	
\node  at (17.25,-3)    ()  {\LARGE Freezing decision};

\draw [->, >=stealth, thick, rounded corners=0.2cm, color = datemagenta]  ( 20.25, -3) -- (21.25,-3) ;	
\node  at (23.75,-3)    ()  {\LARGE Training dynamics};
        
            \end{tikzpicture}

%% file: fig4.tex
\begin{tikzpicture}[scale=.6, transform shape];

\node [draw,fill=gray!30,rotate=0,anchor=north west, minimum width=8cm,minimum height=14cm, rounded corners=.05cm,opacity=.2, dashed]  at ( -1, 11.75)    (box1)          {};

\node [draw,fill=datemagenta!50,rotate=0,anchor=north west, minimum width=3cm,minimum height=0.75cm, rounded corners=.05cm]  at ( 0, 10)    (l1){\LARGE  Layer \#1};				
		
\node [draw,fill=datemagenta!50,rotate=0,anchor=north west, minimum width=3cm,minimum height=0.75cm, rounded corners=.05cm]  at ( 0, 8.5)    (l2){\LARGE  Layer \#2};

\node [draw,fill=dateblue!50,rotate=0,anchor=north west, minimum width=3cm,minimum height=0.75cm, rounded corners=.05cm]  at ( 0, 7)    (l3){\LARGE  Layer \#3};

\node [draw,fill=dateblue!50,rotate=0,anchor=north west, minimum width=3cm,minimum height=0.75cm, rounded corners=.05cm]  at ( 0, 5.5)    (l4){\LARGE  Layer \#4};

\node [draw,fill=datemagenta!50,rotate=0,anchor=north west, minimum width=3cm,minimum height=0.75cm, rounded corners=.05cm]  at ( 0, 4)    (l5){\LARGE  Layer \#5};

\node [draw,fill=dateblue!50,rotate=0,anchor=north west, minimum width=3cm,minimum height=0.75cm, rounded corners=.05cm]  at ( 0, 2.5)    (l6){\LARGE  Layer \#6};

\node [draw,fill=dateblue!50,rotate=0,anchor=north west, minimum width=3cm,minimum height=0.75cm, rounded corners=.05cm]  at ( 0, 1)    (l7){\LARGE  Layer \#7};

\node [draw,fill=datemagenta!50,rotate=0,anchor=north west, minimum width=3cm,minimum height=0.75cm, rounded corners=.05cm]  at ( 0, -0.5)    (l8){\LARGE  Layer \#8};
				
\draw [->, >=stealth, thick, rounded corners=0.2cm] ([yshift=0.75cm] l1.north) -- (l1.north);

\draw [->, >=stealth, thick, rounded corners=0.2cm] (l1.south) -- (l2.north);

\draw [->, >=stealth, thick, rounded corners=0.2cm] (l2.south) -- (l3.north);

\draw [->, >=stealth, thick, rounded corners=0.2cm] (l3.south) -- (l4.north);

\draw [->, >=stealth, thick, rounded corners=0.2cm] (l4.south) -- (l5.north);				

\draw [->, >=stealth, thick, rounded corners=0.2cm] (l5.south) -- (l6.north);

\draw [->, >=stealth, thick, rounded corners=0.2cm] (l6.south) -- (l7.north);

\draw [->, >=stealth, thick, rounded corners=0.2cm] (l7.south) -- (l8.north);

\draw [->, >=stealth, thick, rounded corners=0.2cm] (l8.south) -- ([yshift=-0.75cm] l8.south);

\node [draw,fill=gray!30,rotate=0,anchor=north west, minimum width=1.5cm,minimum height=0.75cm, rounded corners=.05cm]  at ( 4.5, 7.5)    (v1){\LARGE \color{datemagenta} 10.5};				
\node [draw,fill=gray!30,rotate=0,anchor=north west, minimum width=1.5cm,minimum height=0.75cm, rounded corners=.05cm]  at ( 4.5, 6.75)    (){\LARGE \color{datemagenta} 12.3};
\node [draw,fill=gray!30,rotate=0,anchor=north west, minimum width=1.5cm,minimum height=0.75cm, rounded corners=.05cm]  at ( 4.5, 6)    (){\LARGE \color{dateblue} 8.5};
\node [draw,fill=gray!30,rotate=0,anchor=north west, minimum width=1.5cm,minimum height=0.75cm, rounded corners=.05cm]  at ( 4.5, 5.25)    (){\LARGE  \color{dateblue} 9.4};
\node [draw,fill=gray!30,rotate=0,anchor=north west, minimum width=1.5cm,minimum height=0.75cm, rounded corners=.05cm]  at ( 4.5, 4.5)    (){\LARGE  \color{datemagenta} 11.4};
\node [draw,fill=gray!30,rotate=0,anchor=north west, minimum width=1.5cm,minimum height=0.75cm, rounded corners=.05cm]  at ( 4.5, 3.75)    (){\LARGE  \color{dateblue} 3.5};
\node [draw,fill=gray!30,rotate=0,anchor=north west, minimum width=1.5cm,minimum height=0.75cm, rounded corners=.05cm]  at ( 4.5, 3)    (){\LARGE  \color{dateblue} 6.7};
\node [draw,fill=gray!30,rotate=0,anchor=north west, minimum width=1.5cm,minimum height=0.75cm, rounded corners=.05cm]  at ( 4.5, 2.25)    (){\LARGE  \color{datemagenta} 13};

\node  at ([yshift=1.25cm] l1.north)    ()  {\LARGE Transformer Model};
\node  at ([yshift=0.5cm] v1.north)    ()  {\LARGE Distance Vector};

\node [ single arrow,fill=datebrown!50,rotate=0,anchor=base,  rounded corners=.05cm, minimum height=1.5cm]  at ( 8, 4.25)    (){\LARGE  ILS};
\node  at ([yshift=0.5cm, xshift=2cm] box1.south)    ()  {\LARGE Iteration $t-1$};


\node [draw,fill=gray!30,rotate=0,anchor=north west, minimum width=8cm,minimum height=14cm, rounded corners=.05cm,opacity=.2, dashed]  at ( -1+10, 11.75)    (box2)          {};

\node [draw,fill=datemagenta!50,rotate=0,anchor=north west, minimum width=3cm,minimum height=0.75cm, rounded corners=.05cm]  at ( 10, 10)    (l1){\LARGE  Layer \#1};				
		
\node [draw,fill=dateblue!50,rotate=0,anchor=north west, minimum width=3cm,minimum height=0.75cm, rounded corners=.05cm]  at ( 10, 8.5)    (l2){\LARGE  Layer \#2};

\node [draw,fill=datemagenta!50,rotate=0,anchor=north west, minimum width=3cm,minimum height=0.75cm, rounded corners=.05cm]  at ( 10, 7)    (l3){\LARGE  Layer \#3};

\node [draw,fill=datemagenta!50,rotate=0,anchor=north west, minimum width=3cm,minimum height=0.75cm, rounded corners=.05cm]  at ( 10, 5.5)    (l4){\LARGE  Layer \#4};

\node [draw,fill=dateblue!50,rotate=0,anchor=north west, minimum width=3cm,minimum height=0.75cm, rounded corners=.05cm]  at ( 10, 4)    (l5){\LARGE  Layer \#5};

\node [draw,fill=dateblue!50,rotate=0,anchor=north west, minimum width=3cm,minimum height=0.75cm, rounded corners=.05cm]  at ( 10, 2.5)    (l6){\LARGE  Layer \#6};

\node [draw,fill=dateblue!50,rotate=0,anchor=north west, minimum width=3cm,minimum height=0.75cm, rounded corners=.05cm]  at ( 10, 1)    (l7){\LARGE  Layer \#7};

\node [draw,fill=datemagenta!50,rotate=0,anchor=north west, minimum width=3cm,minimum height=0.75cm, rounded corners=.05cm]  at ( 10, -0.5)    (l8){\LARGE  Layer \#8};
				
\draw [->, >=stealth, thick, rounded corners=0.2cm] ([yshift=0.75cm] l1.north) -- (l1.north);

\draw [->, >=stealth, thick, rounded corners=0.2cm] (l1.south) -- (l2.north);

\draw [->, >=stealth, thick, rounded corners=0.2cm] (l2.south) -- (l3.north);

\draw [->, >=stealth, thick, rounded corners=0.2cm] (l3.south) -- (l4.north);

\draw [->, >=stealth, thick, rounded corners=0.2cm] (l4.south) -- (l5.north);				

\draw [->, >=stealth, thick, rounded corners=0.2cm] (l5.south) -- (l6.north);

\draw [->, >=stealth, thick, rounded corners=0.2cm] (l6.south) -- (l7.north);

\draw [->, >=stealth, thick, rounded corners=0.2cm] (l7.south) -- (l8.north);

\draw [->, >=stealth, thick, rounded corners=0.2cm] (l8.south) -- ([yshift=-0.75cm] l8.south);

\node [draw,fill=gray!30,rotate=0,anchor=north west, minimum width=1.5cm,minimum height=0.75cm, rounded corners=.05cm]  at ( 4.5 + 10, 7.5)    (v1){\LARGE \color{datemagenta} 8.1};				
\node [draw,fill=gray!30,rotate=0,anchor=north west, minimum width=1.5cm,minimum height=0.75cm, rounded corners=.05cm]  at ( 4.5 + 10, 6.75)    (){\LARGE \color{dateblue} 5.3};
\node [draw,fill=gray!30,rotate=0,anchor=north west, minimum width=1.5cm,minimum height=0.75cm, rounded corners=.05cm]  at ( 4.5 + 10, 6)    (){\LARGE \color{datemagenta} 8.5};
\node [draw,fill=gray!30,rotate=0,anchor=north west, minimum width=1.5cm,minimum height=0.75cm, rounded corners=.05cm]  at ( 4.5 + 10, 5.25)    (){\LARGE  \color{datemagenta} 9.4};
\node [draw,fill=gray!30,rotate=0,anchor=north west, minimum width=1.5cm,minimum height=0.75cm, rounded corners=.05cm]  at ( 4.5 + 10, 4.5)    (){\LARGE  \color{dateblue} 7.1};
\node [draw,fill=gray!30,rotate=0,anchor=north west, minimum width=1.5cm,minimum height=0.75cm, rounded corners=.05cm]  at ( 4.5 + 10, 3.75)    (){\LARGE  \color{dateblue} 3.5};
\node [draw,fill=gray!30,rotate=0,anchor=north west, minimum width=1.5cm,minimum height=0.75cm, rounded corners=.05cm]  at ( 4.5 + 10, 3)    (){\LARGE  \color{dateblue} 6.7};
\node [draw,fill=gray!30,rotate=0,anchor=north west, minimum width=1.5cm,minimum height=0.75cm, rounded corners=.05cm]  at ( 4.5 + 10, 2.25)    (){\LARGE  \color{datemagenta} 11};

\node  at ([yshift=1.25cm] l1.north)    ()  {\LARGE Transformer Model};
\node  at ([yshift=0.5cm] v1.north)    ()  {\LARGE Distance Vector};

\node [ single arrow,fill=datebrown!50,rotate=0,anchor=base,  rounded corners=.05cm, minimum height=1.5cm]  at ( 18, 4.25)    (){\LARGE  ILS};
\node  at ([yshift=0.5cm, xshift=2cm] box2.south)    ()  {\LARGE Iteration $t$};


\node [draw,fill=gray!30,rotate=0,anchor=north west, minimum width=8cm,minimum height=14cm, rounded corners=.05cm,opacity=.2, dashed]  at ( -1 + 20, 11.75)    (box3)          {};

\node [draw,fill=dateblue!50,rotate=0,anchor=north west, minimum width=3cm,minimum height=0.75cm, rounded corners=.05cm]  at ( 20, 10)    (l1){\LARGE  Layer \#1};				
		
\node [draw,fill=dateblue!50,rotate=0,anchor=north west, minimum width=3cm,minimum height=0.75cm, rounded corners=.05cm]  at ( 20, 8.5)    (l2){\LARGE  Layer \#2};

\node [draw,fill=dateblue!50,rotate=0,anchor=north west, minimum width=3cm,minimum height=0.75cm, rounded corners=.05cm]  at ( 20, 7)    (l3){\LARGE  Layer \#3};

\node [draw,fill=datemagenta!50,rotate=0,anchor=north west, minimum width=3cm,minimum height=0.75cm, rounded corners=.05cm]  at ( 20, 5.5)    (l4){\LARGE  Layer \#4};

\node [draw,fill=datemagenta!50,rotate=0,anchor=north west, minimum width=3cm,minimum height=0.75cm, rounded corners=.05cm]  at ( 20, 4)    (l5){\LARGE  Layer \#5};

\node [draw,fill=dateblue!50,rotate=0,anchor=north west, minimum width=3cm,minimum height=0.75cm, rounded corners=.05cm]  at ( 20, 2.5)    (l6){\LARGE  Layer \#6};

\node [draw,fill=datemagenta!50,rotate=0,anchor=north west, minimum width=3cm,minimum height=0.75cm, rounded corners=.05cm]  at ( 20, 1)    (l7){\LARGE  Layer \#7};

\node [draw,fill=datemagenta!50,rotate=0,anchor=north west, minimum width=3cm,minimum height=0.75cm, rounded corners=.05cm]  at (20, -0.5)    (l8){\LARGE  Layer \#8};
				
\draw [->, >=stealth, thick, rounded corners=0.2cm] ([yshift=0.75cm] l1.north) -- (l1.north);

\draw [->, >=stealth, thick, rounded corners=0.2cm] (l1.south) -- (l2.north);

\draw [->, >=stealth, thick, rounded corners=0.2cm] (l2.south) -- (l3.north);

\draw [->, >=stealth, thick, rounded corners=0.2cm] (l3.south) -- (l4.north);

\draw [->, >=stealth, thick, rounded corners=0.2cm] (l4.south) -- (l5.north);				

\draw [->, >=stealth, thick, rounded corners=0.2cm] (l5.south) -- (l6.north);

\draw [->, >=stealth, thick, rounded corners=0.2cm] (l6.south) -- (l7.north);

\draw [->, >=stealth, thick, rounded corners=0.2cm] (l7.south) -- (l8.north);

\draw [->, >=stealth, thick, rounded corners=0.2cm] (l8.south) -- ([yshift=-0.75cm] l8.south);

\node [draw,fill=gray!30,rotate=0,anchor=north west, minimum width=1.5cm,minimum height=0.75cm, rounded corners=.05cm]  at ( 4.5 + 20, 7.5)    (v1){\LARGE \color{dateblue} 4.2};				
\node [draw,fill=gray!30,rotate=0,anchor=north west, minimum width=1.5cm,minimum height=0.75cm, rounded corners=.05cm]  at ( 4.5 + 20, 6.75)    (){\LARGE \color{dateblue} 5.3};
\node [draw,fill=gray!30,rotate=0,anchor=north west, minimum width=1.5cm,minimum height=0.75cm, rounded corners=.05cm]  at ( 4.5 + 20, 6)    (){\LARGE \color{dateblue} 4.9};
\node [draw,fill=gray!30,rotate=0,anchor=north west, minimum width=1.5cm,minimum height=0.75cm, rounded corners=.05cm]  at ( 4.5 + 20, 5.25)    (){\LARGE  \color{datemagenta} 6.4};
\node [draw,fill=gray!30,rotate=0,anchor=north west, minimum width=1.5cm,minimum height=0.75cm, rounded corners=.05cm]  at ( 4.5 + 20, 4.5)    (){\LARGE  \color{datemagenta} 7.1};
\node [draw,fill=gray!30,rotate=0,anchor=north west, minimum width=1.5cm,minimum height=0.75cm, rounded corners=.05cm]  at ( 4.5 + 20, 3.75)    (){\LARGE  \color{dateblue} 3.5};
\node [draw,fill=gray!30,rotate=0,anchor=north west, minimum width=1.5cm,minimum height=0.75cm, rounded corners=.05cm]  at ( 4.5 + 20, 3)    (){\LARGE  \color{datemagenta} 6.7};
\node [draw,fill=gray!30,rotate=0,anchor=north west, minimum width=1.5cm,minimum height=0.75cm, rounded corners=.05cm]  at ( 4.5 + 20, 2.25)    (){\LARGE  \color{datemagenta} 10};

\node  at ([yshift=1.25cm] l1.north)    ()  {\LARGE Transformer Model};
\node  at ([yshift=0.5cm] v1.north)    ()  {\LARGE Distance Vector};

\node  at ([yshift=0.5cm, xshift=2cm] box3.south)    ()  {\LARGE Iteration $t+1$};

\node [draw,fill=dateblue!50,rotate=0,anchor=north west, minimum width=3cm,minimum height=0.75cm, rounded corners=.05cm]  at ( 0, -3)    (l1){\LARGE  Frozen};	
\node  at ([xshift=5.5cm] l1.east)    ()  {\huge Frozen layers are not being updated.};

\node [draw,fill=datemagenta!50,rotate=0,anchor=north west, minimum width=3cm,minimum height=0.75cm, rounded corners=.05cm]  at ( 15, -3)    (l2){\LARGE  Active};	
\node  at ([xshift=5cm] l2.east)    ()  {\huge Active layers are being updated.};

\end{tikzpicture}

%% file: fig5.tex
\begin{tikzpicture}[scale=.6, transform shape];

\node [draw,fill=dateblue!50,rotate=0,anchor=north west, minimum width=6cm,minimum height=0.75cm, rounded corners=.05cm]  at ( 0, 10)    (l1){\large  Layer \#1 (Size: B * 128 * 768)};				
		
\node [draw,fill=datemagenta!50,rotate=0,anchor=north west, minimum width=6cm,minimum height=0.75cm, rounded corners=.05cm]  at ( 0, 8.5)    (l2){\large  Layer \#2 (Size: B * 128 * 3072)};

\node [draw,fill=dateblue!50,rotate=0,anchor=north west, minimum width=6cm,minimum height=0.75cm, rounded corners=.05cm]  at ( 0, 7)    (l3){\large  Layer \#3 (Size: B * 128 * 768)};

\node [draw,fill=datemagenta!50,rotate=0,anchor=north west, minimum width=6cm,minimum height=0.75cm, rounded corners=.05cm]  at ( 0, 5.5)    (l4){\large  Layer \#4 (Size: B * 128 * 3072)};




				
\draw [->, >=stealth, thick, rounded corners=0.2cm] ([yshift=0.75cm] l1.north) -- (l1.north);

\draw [->, >=stealth, thick, rounded corners=0.2cm] (l1.south) -- (l2.north);

\draw [->, >=stealth, thick, rounded corners=0.2cm] (l2.south) -- (l3.north);

\draw [->, >=stealth, thick, rounded corners=0.2cm] (l3.south) -- (l4.north);





\draw [->, >=stealth, thick, rounded corners=0.2cm] (l4.south) -- ([yshift=-0.75cm] l4.south);

\node  at ([yshift=1.25cm] l1.north)    ()  {\Large Transformer-based Model};

\node [draw,fill=dateblue!50,rotate=0,anchor=north west, minimum width=3cm,minimum height=0.75cm, rounded corners=.05cm]  at ( 8, 11)    (l1){\large  Frozen};

\node [draw,fill=datemagenta!50,rotate=0,anchor=north west, minimum width=3cm,minimum height=0.75cm, rounded corners=.05cm]  at (8, 9)    (l2){\large  Active};	
\node  at ([xshift=-1.5cm, yshift=-2cm] l2.east)    ()  {\color{dateblue} \Large  \textbf{B: Batch Size} };
\node  at ([xshift=-1.5cm, yshift=-4cm] l2.east)    ()  {\color{dateblue} \Large  \textbf{Memory Reduction =} $\mathbf{1.25\times}$};

\end{tikzpicture}

%% file: tab1.tex
\begin{tabular}{ c|c|c|c} 
 \hline
 \hline
 Type of Layer & Description & \# Activations & Status\\ 
 \hline
 Dense & attention.self.query & $B * T * H$  & Balance\\ 
 Dense & attention.self.key & $B * T * H$  & Balance\\
 Dense & attention.self.value & $B * T * H$  & Balance\\
 Dense & attention.output & $B * T * H$  & Balance\\
 LayerNorm & attention.output & $B * T * H$  & Balance\\
 Dense & intermediate & $B * T * H$  & Balance\\
 \textbf{Dense} & \textbf{output} & $\mathbf{B * T * 4 * H}$  & \textbf{Imbalance}\\
 LayerNorm & output & $B * T * H$  & Balance\\
 \hline
 \hline
\end{tabular}

%% file: tab2.tex
\begin{tabular}{ c|c|c||c|c|c} 
 \hline
 \hline
 Type of Layer  & \# Activations & Type of Activations &Type of Layer  & \# Activations & Type of Activations\\ 
 \hline
 Dense  & $B * T * H$  & Dynamic & \textbf{LayerNorm}  & $\mathbf{B * T * H}$  & \textbf{Static}\\
 \textbf{MatMul}  & $\mathbf{B * T * H}$ ($\mathbf{2\times}$) & \textbf{Static} &  Dense  & $B * T * H$  & Dynamic\\
 \textbf{Softmax} & $\mathbf{B * T * T}$ & \textbf{Static} & \textbf{GELU}  & $\mathbf{B * T * 4 * H}$  & \textbf{Static} \\
 \textbf{MatMul}  & $\mathbf{B * T * H}$ \& $\mathbf{B * T * T}$ & \textbf{Static} & Dense  & $B * T * 4 * H$  & Dynamic \\
 Dense  & $B * T * H$  & Dynamic & \textbf{LayerNorm}  & $\mathbf{B * T * H}$  & \textbf{Static}\\

 \hline
 \hline
\end{tabular}

%% file: tab3.tex
\begin{tabular}{ c|c|c|c|c|c|c|c|c|c|c} 
 \hline
 \hline
 Method  & Metric & MNLI$_{m }$ & QQP  & QNLI & SST-2 & CoLA & STS-B & MRPC & RTE & SQuAD 2.0\\ 
 \hline
\multirow{3}{*}{BERT (Basline)} & Accuracy                   & 83.4 & 90.8 & 90.5 & 92.1 & 58.9 & 89.5 & 86.4 & 70.2 & 74.0\\
                                & Memory of Activations (GB) & 3.2 & 3.2 & 3.2 & 3.2 & 3.2 & 3.2 & 3.2 & 3.2 & 55.1\\
                                & Total On-chip GPU Memory (GB)          & 6.1 &6.1 &6.1 &6.1 &6.1 &6.1 &6.1 &6.1 & 58.5 (2 GPUs) \\
\hline                               
\multirow{4}{*}{{\sc SlimFit}       } & Accuracy                   & 83.3 & 90.4 & 90.4 & 92.3 & 59.6 & 89.4 & 86.3 & 70.4 & 74.0\\
                                & Freezing Rate (\%)         & 80 & 80 &  95 & 95 & 90 & 85 & 91 & 90 & 80 \\
                                & Memory of Activations (GB) & 0.7 & 0.7 & 0.5 & 0.5 & 0.6 & 0.7 & 0.6 & 0.6 & 10\\
                                & Total On-chip GPU Memory (GB)          & 4.4 & 4.4 & 4.0 & 4.0 & 4.3 & 4.3 & 4.3 & 4.3 & 19.1\\
                                
 \hline
 \hline
\end{tabular}

%% file: fig7.tex
\begin{tikzpicture}
  \centering
  \begin{axis}[
        ybar, axis on top,
        height=4cm, width=25cm,
        bar width=0.25cm,
        ymajorgrids, tick align=inside,
        major grid style={draw =white},
        enlarge y limits={value=.1,upper},
        ymin=0, ymax=90,
        axis x line*=bottom,
        axis y line*=left,
        y axis line style={opacity=0},
        tickwidth=0pt,
        enlarge x limits={abs = 1.5cm},
        x = 3cm,
        legend style={font=\footnotesize,
            at={(0.5,-0.2)},
            anchor=north,
            legend columns=-1,
            /tikz/every even column/.append style={column sep=0.3cm}
        },
        ylabel style={align=center},
        ylabel={Total On-device\\GPU Memory (GB)},
        symbolic x coords={
         GLUE, SQuAD 2.0, CIFAR-10, CIFAR-100, ImageNet  },
       xtick=data,
       nodes near coords=\rotatebox{90}{
        \pgfmathprintnumber[precision=1]{\pgfplotspointmeta}
       }
    ]
    \addplot [draw=none, fill=Paired-1!60] coordinates {
      (GLUE, 6.1)
      (SQuAD 2.0, 16.4)
      (CIFAR-10,7.2)
      (CIFAR-100,7.2)
      (ImageNet,20.5)};
   \addplot [draw=none,fill=Paired-3!60] coordinates {
      (GLUE, 4.0)
      (SQuAD 2.0, 7.3)
      (CIFAR-10,4.3)
      (CIFAR-100,4.5)
      (ImageNet,8.9)};
   \addplot [draw=none,fill=Paired-5!60] coordinates {
      (GLUE, 9.1)
      (SQuAD 2.0, 29.6)
      (CIFAR-10,11.4)
      (CIFAR-100,11.4)
      (ImageNet,40.1)};
    \addplot [draw=none,fill=Paired-7!60] coordinates {
      (GLUE, 5.1)
      (SQuAD 2.0, 11.1)
      (CIFAR-10,5.8)
      (CIFAR-100,6.3)
      (ImageNet,14.9)};
   \addplot [draw=none,fill=Paired-9!60] coordinates {
      (GLUE, 15.4)
      (SQuAD 2.0, 58.5)
      (CIFAR-10,20.3)
      (CIFAR-100,20.3)
      (ImageNet,77.4)};   
    \addplot [draw=none,fill=Paired-11!60] coordinates {
      (GLUE, 6.8)
      (SQuAD 2.0, 19.1)
      (CIFAR-10,8.7)
      (CIFAR-100,9.6)
      (ImageNet,26.1)};

    \legend{Baseline-32, {\sc SlimFit}-32, Baseline-64, {\sc SlimFit}-64, Baseline-128, {\sc SlimFit}-128}
  \end{axis}
  \end{tikzpicture}

%% file: tab4_1.tex
\begin{tabular}{ c|c|c|c|c||c|c|c} 
 \hline
 \hline
  & & \multicolumn{3}{c||}{Baseline} & \multicolumn{3}{c}{{\sc SlimFit}}\\
 \hline
 Model  & Metric & CIFAR-10 & CIFAR-100 & ImageNet & CIFAR-10 & CIFAR-100 & ImageNet\\ 
 \hline
\multirow{4}{*}{ViT} & Accuracy (\%)                              & 98.8 & 91.2 & 83.3 & 98.5 & 91.0 & 83.3\\
                                & Freezing Rate (\%)         & NA & NA & NA & 90 & 75 & 95\\
                                & Memory of Activations (GB) & 4.5 & 4.5 & 69.5 & 0.8 & 1.0 & 11.9\\
                                & Total Memory (GB)          & 7.2 & 7.2 & 77.4 (3 GPUs) & 4.3 & 4.5 & 26.1\\         
    
 \hline
 \hline
\end{tabular}

%% file: fig6a.tex
\pgfplotsset{width=10cm, height = 5cm}
\begin{tikzpicture}
\begin{axis}[
    xlabel={Freezing Rate (\%)},
    ymin=40,
    ymax=65,
    xmin = 0,
    xmax = 100,
    ylabel=Matthew's Correlation,
    legend pos=south west,
    label style={font=\large},
    legend style={font=\footnotesize},
    legend entries={ILS, Random, Progressive},
    ]
    \addplot[color=Paired-1, thick, smooth,each nth point={1}, mark=star] %
    table[x=i,
          y=x,
          col sep=space]
          {./cola_acc.txt};
     \addplot[color=Paired-5, thick, smooth,each nth point={1}, mark=o] %
    table[x=i,
          y=y,
          col sep=space]
          {./cola_acc.txt};
     \addplot[color=Paired-3, thick, smooth,each nth point={1}, mark=triangle] %
    table[x=j,
          y=z,
          col sep=space]
          {./cola_acc.txt};
\end{axis}
\end{tikzpicture}

%% file: fig6b.tex
\pgfplotsset{width=10cm, height = 5cm}
\begin{tikzpicture}
\begin{axis}[
    xlabel={Freezing Rate (\%)},
    ymin=60,
    ymax=100,
    xmin = 0,
    xmax = 100,
    ylabel=Accuracy (\%),
    legend pos=south west,
    label style={font=\large},
    legend style={font=\footnotesize},
    legend entries={ILS, Random, Progressive},
    ]
    \addplot[color=Paired-1, thick, smooth,each nth point={1}, mark=star] %
    table[x=i,
          y=x,
          col sep=space]
          {./mrpc_acc.txt};
     \addplot[color=Paired-5, thick, smooth,each nth point={1}, mark=o] %
    table[x=i,
          y=y,
          col sep=space]
          {./mrpc_acc.txt};
     \addplot[color=Paired-3, thick, smooth,each nth point={1}, mark=triangle] %
    table[x=j,
          y=z,
          col sep=space]
          {./mrpc_acc.txt};
\end{axis}
\end{tikzpicture}

%% file: tab6_1.tex
\begin{tabular}{ c|c|c|c|c|c} 
 \hline
 \hline
  Model  & Metric & Baseline & 4-bit GACT (ICML'22 \cite{GACT}) & DropIT (ICLR'23 \cite{DropIT}) & {\sc SlimFit}\\
 \hline
\multirow{5}{*}{BERT} & Accuracy (Matthew's Correlation)     & 58.9 & 59.0 & 57.5 & \textbf{59.6}\\
                                & Freezing Rate (\%)         & NA & NA & NA & 90\% \\
                                & Memory of Activations (GB) & 3.2 & \textbf{0.5} & 2.4 & 0.6\\
                                & Total Memory (GB)          & 6.1 & 6.0 & 5.7 & \textbf{4.3} \\
                                & Latency (Seconds)          & \textbf{251} & 455 & 367 & 281\\
 \hline
 \hline
\end{tabular}

%% file: fig9a.tex
\pgfplotsset{width=7.5cm, height = 3.5cm}
\begin{tikzpicture}
\begin{axis}[
    xlabel={Training Iteration},
    ymin=65,
    ymax=75,
    xmin = 0,
    xmax = 15,
    ylabel style={align=center},
    ylabel=Memory of\\Activations (GB),
    grid = major,
    legend pos=north east,
    label style={font=\footnotesize},
    legend style={font=\footnotesize},
    legend entries={Baseline},
    ]
    \addplot[color=Paired-1, thick, each nth point={1}, mark=star] %
    table[x=i,
          y=y,
          col sep=space]
          {./MEM.txt};
\end{axis}
\end{tikzpicture}
\begin{tikzpicture}
\begin{axis}[
    xlabel={Training Iteration},
    ymin=11, 
    ymax=13, 
    xmin = 0,
    xmax = 15,
    ylabel style={align=center},
    ylabel=Memory of\\Activations (GB),
    grid = major,
    legend pos=north east,
    label style={font=\footnotesize},
    legend style={font=\footnotesize},
    legend entries={SlimFit},
    ]
    \addplot[color=Paired-5, thick, each nth point={1}, mark=*] %
    table[x=i,
          y=t,
          col sep=space]
          {./MEM.txt};

\end{axis}

\end{tikzpicture}

%% file: fig9b.tex
\pgfplotsset{width=7.5cm, height = 3.5cm}
\begin{tikzpicture}
\begin{axis}[
    xlabel={Training Iteration},
    ymin=0,
    ymax=90,
    xmin = 0,
    xmax = 15,
    ylabel style={align=center},
    ylabel=Total On-device\\memory (GB),
    grid = major,
    legend pos=south east,
    label style={font=\footnotesize},
    legend style={font=\footnotesize},
    legend entries={Baseline},
    ]
    \addplot[color=Paired-1, thick, each nth point={1}, mark=star] %
    table[x=i,
          y=x,
          col sep=space]
          {./MEM.txt};
\end{axis}
\end{tikzpicture}
\begin{tikzpicture}
\begin{axis}[
    xlabel={Training Iteration},
    ymin=0,
    ymax=30,
    xmin = 0,
    xmax = 15,
    ylabel style={align=center},
    ylabel=Total On-device\\memory (GB),
    grid = major,
    legend pos=south east,
    label style={font=\footnotesize},
    legend style={font=\footnotesize},
    legend entries={SlimFit},
    ]
    \addplot[color=Paired-5, thick, each nth point={1}, mark=*] %
    table[x=i,
          y=z,
          col sep=space]
          {./MEM.txt};
\end{axis}
\end{tikzpicture}

%% file: fig2.tex
%
%
%
%
%
%

\begin{tikzpicture}[scale=.6, transform shape];
				
				\node [draw,fill=gray!30,rotate=0,anchor=north west, minimum width=7.75cm,minimum height=7cm, rounded corners=.05cm,opacity=.5]  at ( 6.75, 7.5)    (box)          {};
				
				\node [draw,fill=Paired-7!30,rotate=0,anchor=north west, minimum width=4.5cm,minimum height=1cm, rounded corners=.05cm,]  at ( 8,     0)    (femb)          {\large Embedding};
				\node [draw,fill=Paired-1!30,rotate=0,anchor=north west, minimum width=4.5cm,minimum height=1cm, rounded corners=.05cm,]  at ( 8,     3)    (fmha)          {\large Multi-head attention};
				\node [draw,fill=Paired-3!10,rotate=0,anchor=north west, minimum width=4.5cm,minimum height=0.5cm, rounded corners=.05cm,]  at ( 8,     4)    (fadd1)          {\large Add \& Layernorm};
				\node [draw,fill=Paired-11!30,rotate=0,anchor=north west, minimum width=4.5cm,minimum height=1cm, rounded corners=.05cm,]  at ( 8,     6)    (ffnn)          {\large Feed-forward network};
				\node [draw,fill=Paired-3!10,rotate=0,anchor=north west, minimum width=4.5cm,minimum height=0.5cm, rounded corners=.05cm,]  at ( 8,     7)    (fadd2)          {\large Add \& Layernorm};

				\draw [->, >=stealth, thick, rounded corners=0.2cm] (femb.north) -- (fmha.south);
				\draw [->, >=stealth, thick, rounded corners=0.2cm] (fmha.north) -- (fadd1.south);
				\draw [->, >=stealth, thick, rounded corners=0.2cm] (fadd1.north) -- (ffnn.south);
				\draw [->, >=stealth, thick, rounded corners=0.2cm] (ffnn.north) -- (fadd2.south);

				\draw [->, >=stealth, thick, rounded corners=0.2cm] (femb.north)+(0,1.25cm) -| ([xshift=-0.75cm] fadd1.west) -- (fadd1.west);
				
				\draw [->, >=stealth, thick, rounded corners=0.2cm] (fadd1.north)+(0,0.5cm) -| ([xshift=-0.75cm] fadd2.west) -- (fadd2.west);
				\draw [->, >=stealth, thick, rounded corners=0.2cm] (fadd2.north) -- ([yshift=1cm] fadd2.north);

				\draw [pen colour={datemagenta!100},decorate, ultra thick,
    				decoration = {calligraphic brace,
    							mirror,
        						  raise=5pt,
        						  amplitude=5pt}] (fmha.south east) --  (fadd1.north east) node[pos=0.5,right=15pt,black]{\Large ${MHA}$};
				
				\draw [pen colour={datemagenta!100},decorate, ultra thick,
    				decoration = {calligraphic brace,
    							mirror,
        						  raise=5pt,
        						  amplitude=5pt}] (ffnn.south east) --  (fadd2.north east) node[pos=0.5,right=15pt,black]{\Large ${FFN}$};
    			
			\node  at ( 14, 0.75)    (heads)  {$\times L$};

            \end{tikzpicture}


%% file: tab8.tex
\begin{tabular}{ c|c|c|c} 
 \hline
 \hline
Module  & Type of Layer & Description & \# Activations \\ 
 \hline
\multirow{8}{*}{MHA} & Dense & attention.query & $B * T * H$ \\
                     & Dense & attention.key & $B * T * H$ \\
                     & Dense & attention.value & $B * T * H$ \\
                     & MatMul & NA & $B * T * H (2\times)$ \\
                     & Softmax & NA & $B * T * T$ \\
                     & MatMul & NA &  $B * T * H$ \& $B * T * T$\\
                     & Dense & attention.output & $B * T * H$ \\
                     & LayerNorm & attention.output & $B * T * H$ \\
\hline
\multirow{4}{*}{FFN} & Dense & intermediate & $B * T * H$ \\
                     & GELU & NA & $B * T * 4 * H$ \\
                     & Dense & output & $B * T * 4 * H$ \\
                     & LayerNorm & output &  $B * T * H$\\
 \hline
 \hline
\end{tabular}

%% file: alg2.tex
\begin{algorithmic}
   \STATE {\bfseries Description:} number of integer bits as $ib$, number of fractional bits as $fb$, input $x$, output $y$
   \vspace{0.25cm}
   \STATE {\bfseries  32 bits to 8 bits conversion:}
   \vspace{0.25cm}
   \STATE $y$ = clamp(round($x*2^{fb}$),$-2^{fb+ib-1}$, $2^{fb+ib-1}-1$)
   \vspace{0.25cm}
   \STATE {\bfseries  8 bits to 32 bits conversion:}
   \vspace{0.25cm}
   \STATE $x$ = $\dfrac{y}{2^{fb}}$
\end{algorithmic}

%% file: alg5.tex
\begin{algorithmic}
   \STATE {\bfseries Description:} $\textbf{x}$: input vector, $\textbf{x}_s$: sorted input vector, $\textbf{x}_{idx}$: indices of the sorted input vector, $\textbf{y}_s$: top $10\%$ largest values, $\textbf{y}_{idx}$: indices of top $10\%$ largest values, $\textbf{y}$: output vector, sort: sorting function, zeros: function to create zero-valued tensor, and scatter: function to replace zero values with the tensor values from $\textbf{y}_s$ according to the indices.
   \vspace{0.25cm}
   \STATE {\bfseries  Pruning process:}
   \vspace{0.25cm}
   \STATE $\textbf{x}_s, \textbf{x}_{idx}$ = sort(\textbf{x})

   \STATE $\textbf{y}_s, \textbf{y}_{idx}$ = $\textbf{x}_s[0:int(\textbf{x}.numel() * 0.1)], \textbf{x}_{idx}[0:int(\textbf{x}.numel() * 0.1)]$
   \vspace{0.25cm}
   \STATE {\bfseries  Restoring process:}
    \vspace{0.25cm}
    \STATE $\textbf{y}$ = zeros($\textbf{x}.numel()$)

    \STATE $\textbf{y}$ = scatter($\textbf{y}, \textbf{y}_s, \textbf{y}_{idx}$)
   
\end{algorithmic}

%% file: tab7.tex
\begin{tabular}{ c|c|c|c|c|c} 
 \hline
 \hline
Dataset  & Model & Optimizer & Learning Rate & Warmup & Evaluation Metric \\ 
 \hline
MNLI$_m$ & bert-base-cased & AdamW & 4e-5 & 0 & percentage accuracy \\
QQP      & bert-base-cased & AdamW & 5e-5 & 0 & percentage accuracy \\
QNLI     & bert-base-cased & AdamW & 5e-5 & 0 & percentage accuracy \\   
SST-2    & bert-base-cased & AdamW & 8e-5 & 0 & percentage accuracy \\
CoLA     & bert-base-cased & AdamW & 8e-5 & 0.1 & Matthew’s correlation \\
STS-B    & bert-base-cased & AdamW & 8e-5 & 0 & Spearman correlation \\
MRPC     & bert-base-cased & AdamW & 1.25e-4& 0 & percentage accuracy \\
RTE      & bert-base-cased & AdamW & 1.2e-4 & 0 & percentage accuracy \\
SQuAD 2.0& bert-base-uncased & AdamW & 1.8e-4 & 0.1 & F1 score \\
CIFAR-10 & vit-base-patch16-224-in21k & AdamW & 7.5e-5 & 0 & percentage accuracy \\
CIFAR-100& vit-base-patch16-224-in21k & AdamW & 5.5e-5 & 0 & percentage accuracy \\
ImageNet & vit-base-patch16-384 & AdamW & 5e-5 & 0 & percentage accuracy \\
 \hline
 \hline
\end{tabular}

%% file: tab5.tex
\begin{tabular}{ c|c|c|c|c|c|c} 
 \hline
 \hline
 \multirow{2}{*}{Dataset}  & \multirow{2}{*}{Baseline} & \multicolumn{3}{c|}{Quantization of} & Pruning of & All \\ 
   & &  Linear & MatMul & GELU & LayerNorm & together \\ 
 \hline
CoLA & 58.9 & 58.9 & 60.6 & 60.0 & 59.7 & 59.7 \\
MRPC & 86.4 & 86.4 & 86.3 & 86.3 & 86.3 & 86.3 \\                           
 \hline
 \hline
\end{tabular}

%% file: alg4.tex
\begin{algorithmic}
   \STATE \textbf{class} ILSFunction(torch.autograd.Function):
   \STATE \hskip1.5em @staticmethod
   \STATE \hskip1.5em \textbf{def} forward(ctx, input: torch.Tensor, weight: torch.nn.Parameter, requires\_grad):
   \STATE \hskip3em \# Compute forward computations to obtain out
   \STATE \hskip3em \textbf{if} requires\_grad:
   \STATE \hskip4.5em ctx.save\_for\_backward(compress(input), weight)
   \STATE \hskip3em \textbf{else}:
   \STATE \hskip4.5em ctx.save\_for\_backward(weight)
   \STATE \hskip3em ctx.requires\_grad = requires\_grad
   \STATE \hskip3em \textbf{return} out
   \STATE \hskip1.5em @staticmethod
   \STATE \hskip1.5em \textbf{def} backward(ctx, grad\_output: torch.Tensor):
   \STATE \hskip3em \textbf{if} ctx.requires\_grad:
   \STATE \hskip4.5em input, weight = decompress(ctx.saved\_tensors[0]), ctx.saved\_tensors[1]
   \STATE \hskip4.5em \# Compute backward computations to obtain grad\_input and grad\_weight
   \STATE \hskip3em \textbf{else}:
   \STATE \hskip4.5em weight = ctx.saved\_tensors[1]
   \STATE \hskip4.5em grad\_weight = None
   \STATE \hskip4.5em \# Compute backward computations to obtain grad\_input
   \STATE \hskip3em \textbf{return} grad\_input, grad\_weight, None
\end{algorithmic}

%% file: fig8.tex
\pgfplotsset{width=10cm, height = 5.5cm}
\begin{tikzpicture}
\begin{axis}[
    xlabel={Freezing Rate (\%)},
    ymin=17,
    ymax=24,
    xmin = 0,
    xmax = 100,
    ylabel=Hours per epoch,
    grid = major,
    legend pos=north east,
    label style={font=\large},
    legend style={font=\footnotesize},
    legend entries={SlimFit},
    ]
    \addplot[color=Paired-5, thick, each nth point={1}, mark=star] %
    table[x=i,
          y=x,
          col sep=space]
          {./WCT.txt};
\end{axis}
\end{tikzpicture}

%% file: alg3.tex
\begin{algorithmic}
\STATE bert.embeddings.word\_embeddings.weight
\STATE bert.embeddings.position\_embeddings.weight
\STATE bert.embeddings.token\_type\_embeddings.weight
\STATE bert.embeddings.LayerNorm.weight
   \FOR{$i = 0$ to 11:}
   \STATE bert.encoder.layer[i].attention.self.query.weight
   \STATE bert.encoder.layer[i].attention.self.key.weight
   \STATE bert.encoder.layer[i].attention.self.value.weight
   \STATE bert.encoder.layer[i].attention.output.dense.weight
   \STATE bert.encoder.layer[i].attention.output.LayerNorm.weight
   \STATE bert.encoder.layer[i].intermediate.dense.weight
   \STATE bert.encoder.layer[i].output.dense.weight
   \STATE bert.encoder.layer[i].output.LayerNorm.weight
   \ENDFOR
   \STATE bert.pooler.dense.weight
   \STATE classifier.weight
\end{algorithmic}